\pdfoutput=1
\documentclass[sigconf]{acmart}

\setcopyright{none}
\AtBeginDocument{%
  }

\usepackage{pifont}

\usepackage{booktabs}
\usepackage{multirow}
\usepackage{array}
\usepackage{url}
\usepackage[table]{xcolor}
\usepackage{colortbl}
\usepackage{makecell}
\usepackage{algorithm}
\usepackage{algpseudocode}
\usepackage[most]{tcolorbox}
\usepackage{enumitem}
\usepackage{placeins}
\usepackage{titletoc}

\definecolor{TemplLCP}{HTML}{E7EDF7}
\definecolor{TemplTR}{HTML}{FBE5D6}
\definecolor{TemplLTM}{HTML}{DCEFE8}
\definecolor{TemplSTR}{HTML}{E8DEF3}

\newtcolorbox{templatebox}[2][]{%
    enhanced, breakable,
    colback=#2,
    colframe=#2!50!black,
    coltitle=white,
    fonttitle=\bfseries\small,
    title={#1},
    boxrule=0.5pt,
    arc=2pt,
    left=5pt, right=5pt, top=3pt, bottom=3pt,
}

\definecolor{SysPromptFrame}{HTML}{4A5568}
\definecolor{SysPromptBack}{HTML}{F7FAFC}
\definecolor{SysPromptTitle}{HTML}{2D3748}

\newtcolorbox{systempromptbox}[1][]{%
    enhanced, breakable,
    colback=SysPromptBack,
    colframe=SysPromptFrame,
    colbacktitle=SysPromptTitle,
    coltitle=white,
    fonttitle=\bfseries\small,
    title={#1},
    boxrule=0.6pt,
    arc=2pt,
    left=6pt, right=6pt, top=4pt, bottom=4pt,
}

\definecolor{LCPColor}{HTML}{D9ECFF}   
\definecolor{LTMColor}{HTML}{DCEFE8}   
\definecolor{TAColor}{HTML}{FAE4D5}    
\definecolor{STRColor}{HTML}{E8DEF3}   
\definecolor{GroupGray}{gray}{0.96}    

\newcommand{\bestLCP}[1]{\cellcolor{LCPColor!65}{\textbf{#1}}}
\newcommand{\bestLTM}[1]{\cellcolor{LTMColor!65}{\textbf{#1}}}
\newcommand{\bestTA}[1]{\cellcolor{TAColor!65}{\textbf{#1}}}
\newcommand{\bestSTR}[1]{\cellcolor{STRColor!65}{\textbf{#1}}}

\definecolor{OAColor}{HTML}{F8D7DA}
\newcommand{\bestOA}[1]{\cellcolor{OAColor}{\textbf{#1}}}

\acmSubmissionID{5157}

\begin{document}

\title{GeoChrono: Benchmarking and Rethinking Long-Term Temporal Understanding in Remote Sensing}

\author{Yujie Li}
\affiliation{%
  \institution{Beijing University of Posts and Telecommunications}
  \city{Haidian}
  \state{Beijing}
  \country{China}
}
\email{liyujie2003@bupt.edu.cn}

\author{Jiancheng Pan}
\affiliation{%
  \institution{Tsinghua University}
  \city{Haidian}
  \state{Beijing}
  \country{China}}
\email{jiancheng.pan.plus@gmail.com}

\author{Zhiwei Wei}
\affiliation{%
  \institution{Hunan Normal University}
  \city{Changsha}
  \state{Hunan}
  \country{China}}
\email{trentonwei@whu.edu.cn}

\author{Jiuniu Wang}
\affiliation{%
 \institution{City University of Hong Kong}
 \city{Hong Kong}
 \country{China}}
\email{wangjiuniu@gmail.com}

\author{Mugen Peng}
\affiliation{%
  \institution{Beijing University of Posts and Telecommunications}
  \city{Haidian}
  \state{Beijing}
  \country{China}}
\email{pmg@bupt.edu.cn}

\author{Wenjia Xu}
\authornote{Corresponding author.}
\affiliation{%
  \institution{Beijing University of Posts and Telecommunications}
  \city{Haidian}
  \state{Beijing}
  \country{China}}
\email{xuwenjia@bupt.edu.cn}

\renewcommand{\shortauthors}{Trovato et al.}

\begin{abstract}
Remote sensing offers an unparalleled vantage point for observing the Earth's long-term surface evolution, yet it demands that a model not only perceive land cover at isolated moments, but also track changes, memorize evolution histories, and reason across time and space. However, existing studies lack a systematic evaluation that dissects these distinct competencies. To fill this gap, we introduce ChronoBench, a multidimensional benchmark that decomposes this task into four progressive cognitive levels (i.e., Land Cover Perception, Temporal Recognition, Long-Term Memory, and Spatio-Temporal Reasoning). The ChronoBench comprises 12 sub-tasks and 17,689 rigorously validated QA (Question-Answer) pairs. Extensive evaluations reveal that mainstream MLLMs fall drastically behind human experts, with Long-Term Memory emerging as the most critical bottleneck. Motivated by this finding, we further propose GeoChrono, an MLLM with enhanced capabilities for tracing, memorizing, and reasoning about long-term geographic evolution. Leveraging the physical prior that geographic parcels remain spatially fixed while their semantics evolve, we design a Temporal Trajectory Encoder~(TempEnc) that constructs per-location temporal trajectories for dedicated land cover evolution modeling, and we introduce a Coarse-to-Fine Token Compressor~(C2FComp) that adaptively preserves dynamic regions while compressing the static background. To support training, we also construct ChronoInstruct, a 104K-sample instruction-tuning dataset spanning all competency levels for training. GeoChrono achieves state-of-the-art performance on ChronoBench, surpassing the leading commercial MLLMs by over 20\%, while C2FComp reduces visual tokens by over 56\% while retaining GeoChrono's 94.6\% performance. The code and data will be available at \url{https://github.com/IntelliSensing/GeoChrono}.
\end{abstract}

\ccsdesc[500]{Computing methodologies~Scene understanding}

\keywords{Multimodal large language model, remote sensing, temporal understanding}
\begin{teaserfigure}
  \includegraphics[width=\textwidth]{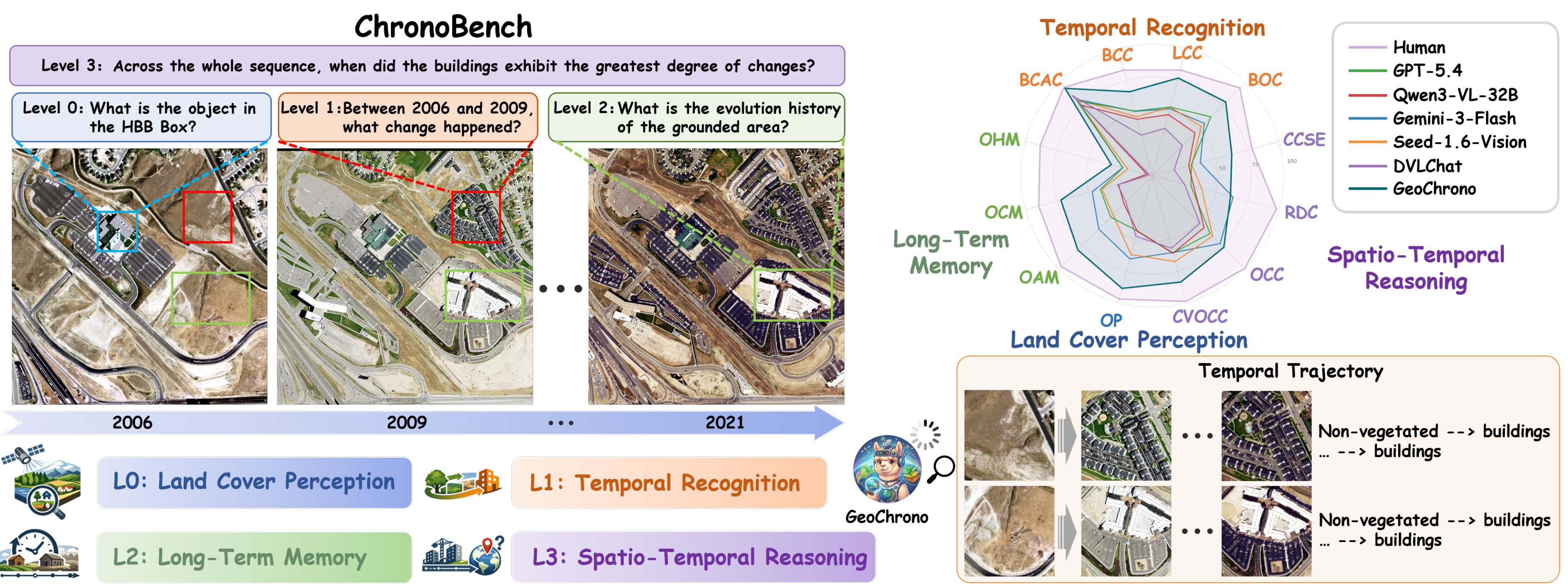}
  \caption{We propose ChronoBench, a multi-dimensional benchmark decomposing long-term remote sensing understanding into a four-level cognitive hierarchy. Built upon this foundation, GeoChrono constructs per-location temporal trajectories to model the long-term evolution of geographic parcels, enabling powerful long-term remote sensing comprehension. GeoChrono achieves state-of-the-art performance on ChronoBench, reaching human-level accuracy on several sub-tasks.}
  \label{fig:teaser}
\end{teaserfigure}


\maketitle

\section{Introduction}
The Earth is a dynamic system evolving along two fundamental axes: space and time. Remote sensing provides an unparalleled vantage point from which to observe this evolution, yet the overwhelming majority of existing interpretation methods confine themselves to \textbf{i)} static snapshots, i.e., understanding what exists at a single moment~\cite{luo2024skysensegpt, pang2025vhm, zhao2026frequency}; or \textbf{ii)} bi-temporal comparisons that capture change between two discrete time points~\cite{liu2024rscama, wang2025disasterm3}. This leaves a critical dimension largely untapped: the continuous, long-term temporal trajectory of landscape transformation, urban expansion, and ecosystem shifts. Meanwhile, multimodal large language models~(MLLMs) have demonstrated remarkable capabilities in long video understanding~\cite{song2024moviechat, weng2024longvlm, shu2025video}, exhibiting an increasingly promising capacity to comprehend extended temporal contexts and reason over complex visual narratives. 
However, long-term remote sensing image sequences differ fundamentally from natural video: geographic parcels remain spatially fixed while their semantics evolve over time, and wide-area scenes contain substantial spatial redundancy because only a small fraction of regions change meaningfully across frames, as illustrated in Figure~\ref{fig:teaser}.
These unique characteristics call for dedicated investigation.

Recent efforts have explored the temporal interpretation of MLLMs in remote sensing through dedicated benchmarks. TEOChatlas~\cite{irvin2024teochat} curates the first temporal remote sensing dataset covering tasks such as temporal scene classification, damage assessment, and change detection, but is constrained by low-resolution imagery~(224 $\times$ 224) and short sequences~(maximum length of 8). DynamicVL~\cite{xuan2025dynamicvl} constructs a high-resolution multi-temporal data suite spanning 42 U.S. cities with 6 urban understanding tasks, revealing that mainstream MLLMs fail in long-term remote sensing understanding. Nevertheless, its task taxonomy is organized around specific application scenarios rather than dissecting the underlying cognitive competencies that models must possess for long-term temporal understanding. While both works reveal that existing MLLMs struggle in long-term remote sensing contexts, neither establishes a structured competency framework that diagnoses \textit{where} and \textit{why} these models fall short. This prompts us to rethink: 
\textit{What competencies are missing in current MLLMs for long-term remote sensing understanding, and how can we bridge these gaps?}

To rigorously investigate this question, we decompose long-term remote sensing understanding into a four-level competency hierarchy, as shown in Figure~\ref{fig:teaser}. \textbf{Level 0 Land Cover Perception}: what is here at a given time? \textbf{Level 1  Temporal Recognition}: what has changed between two time points? \textbf{Level 2 Long-Term Memory}: what was here before, and what transitions have occurred throughout the observation period? \textbf{Level 3 Spatio-Temporal Reasoning}: given all perceived changes and memorized histories, what spatio-temporal relationships can be inferred? Inspired by human cognition~\cite{Tversky_2008_cognition}, these four levels form a progressive hierarchy from basic perception, through change recognition and historical change events memorization, to high-level reasoning, together constituting the cognitive foundation for comprehensive long-term remote sensing understanding.

To this end, we introduce ChronoBench, a comprehensive, multi-dimensional, and multi-granularity benchmark for high-resolution long-temporal remote sensing understanding. ChronoBench is organized around the four competencies outlined above, encompassing 12 sub-tasks and 17,689 rigorously validated QA pairs derived from 3,469 high-resolution images~(1024$\times$1024) spanning 500 distinct regions across 39 major U.S. cities, selected for their rich and diverse land-cover dynamics over extended time spans. During construction, we first design structured question templates for each sub-task, then employ deterministic rule-based programs to traverse pre-existing manually annotated semantic change masks and automatically populate the templates at scale, followed by rigorous manual verification to ensure quality. We evaluate a broad spectrum of mainstream MLLMs, e.g., GPT-5~\cite{openai2025gpt41}, Gemini-3~\cite{team2023gemini}, Qwen3-VL~\cite{bai2025qwen3}, on ChronoBench, and additionally recruit three domain experts to establish a human-level baseline. The results reveal a substantial performance gap between current models and human-level understanding, particularly on tasks related to tracking long-term evolution histories, where existing models exhibit pronounced deficiencies in memorizing and retrieving temporal trajectories of geographic entities.

Motivated by these findings, we develop GeoChrono, an MLLM that enhances tracing, memorization, and reasoning over the long-term evolution of geographic entities. Leveraging the physical prior that each geographic parcel remains spatially fixed while its semantics evolve, we design a \textbf{Temp}oral Trajectory \textbf{Enc}oder~(TempEnc) that explicitly constructs per-location temporal evolution trajectories, strengthening the model's capacity for perceiving temporal changes, memorizing evolution histories, and reasoning across time. Furthermore, to alleviate the substantial computational overhead imposed by high-resolution long-temporal sequences, we exploit the spatial redundancy inherent in wide-area remote sensing scenes and introduce a \textbf{C}oarse-to-\textbf{F}ine Token \textbf{Comp}ressor~(C2FComp). C2FComp leverages prompt text embeddings to assess the task relevance of each spatial region, selectively preserving full-resolution fine tokens for salient areas while condensing the static background into compact coarse representations, thereby drastically compressing the visual token sequence input to the LLM. To support training across all competency dimensions, we further construct ChronoInstruct, a large-scale instruction-tuning dataset containing over 104K samples. Extensive experiments on ChronoBench demonstrate that GeoChrono outperforms previous MLLMs by a substantial margin, achieving an overall accuracy of 78.34\%. Moreover, the C2FComp reduces the visual token count by over 56\% while retaining 94.6\% of the full model's performance~(74.11\% vs. 78.34\%), demonstrating that instruction-guided selective compression can effectively reconcile computational efficiency with long-temporal understanding capability.
The contributions of our work are as follows:
\begin{itemize}
    \item We establish a comprehensive data foundation for long-term remote sensing understanding. Specifically, we propose a four-level cognitive hierarchy, from land cover perception through temporal recognition and long-term memory to spatio-temporal reasoning, and instantiate it as ChronoBench, a multi-dimensional benchmark comprising 12 sub-tasks and 17,689 QA pairs. We further construct ChronoInstruct, a large-scale instruction-tuning dataset comprising over 104K samples spanning all competency levels to support advanced model development.
    \item We design a Temporal Trajectory Encoder~(TempEnc) module that leverages the geostationary prior in remote sensing imagery to decouple the spatio-temporal feature volume into per-location temporal trajectories, thereby enabling dedicated long-term temporal modeling in remote sensing.
    \item We propose a Coarse-to-Fine Token Compressor~(C2FComp) that leverages the spatial sparsity of temporal change to selectively compress visual token sequences under the guidance of prompt text embeddings, thereby substantially balancing the computational overhead of long temporal sequences and model performance.
    \item Extensive experiments on ChronoBench demonstrate that GeoChrono surpasses the leading commercial MLLMs by over 20\%, significantly advancing capabilities across all four competency dimensions of long-term remote sensing understanding. Furthermore, C2FComp reduces the visual token count by over 56\% while retaining 94.6\% of the full model's performance, validating the effectiveness of prompt-guided selective compression.
\end{itemize}

\section{Related Work}
\subsection{Multimodal Large Language Models for RS}
Multimodal large language models~(MLLMs)~\cite{li2023blip, alayrac2022flamingo,liu2023visual, liu2024improved} have achieved remarkable performance across a wide range of vision-language understanding tasks~\cite{wang2025lvbench, zou2025hlv, fu2025video, chandrasegaran2024hourvideo, wu2023multimodal, yin2024survey}. In remote sensing, representative works such as GeoChat~\cite{kuckreja2024geochat}, EarthGPT~\cite{zhang2024earthgpt}, and LHRS-Bot~\cite{muhtar2024lhrs} have been developed for various static interpretation tasks~\cite{zhang2024review, thapa2023deep, lobry2020rsvqa, zhan2023rsvg, wang2025geollava}, yet they predominantly focus on spatial understanding of single-time imagery, leaving temporal evolution largely underexplored. Recent studies have begun to address temporal aspects of remote sensing. ChangeChat~\cite{deng2025changechat} and BTCChat~\cite{li2025btcchat} target bi-temporal change comprehension. Furthermore, TEOChat~\cite{irvin2024teochat}, EarthDial~\cite{soni2025earthdial}, and DVLChat~\cite{xuan2025dynamicvl} further extend this capability to longer temporal sequences by curating multi-temporal instruction-following datasets and training MLLMs accordingly. Nevertheless, these models are developed and trained around specific pre-defined temporal tasks, such as temporal scene classification~\cite{pelletier2019temporal}, building damage assessment~\cite{wang2025disasterm3}, and change detection~\cite{liu2022remote}, without systematically analyzing and addressing the intrinsic challenges that long-temporal remote sensing understanding itself poses. In this work, we revisit this problem and propose GeoChrono, a dedicated temporal trajectory modeling approach for long-term earth observation.


\subsection{Multimodal Benchmarks in Remote Sensing}
The development of remote sensing MLLMs has been accompanied by a growing ecosystem of benchmarks, spanning single-image understanding~\cite{hu2025rsgpt, ou2025geopix, shabbir2025geopixel, wang2025xlrs, zhou2026dvgbench} and bi-temporal change reasoning~\cite{yuan2022change, wang2025disasterm3}, yet extension to longer temporal sequences remains nascent. TEOChatlas~\cite{irvin2024teochat} assembles samples from multiple EO datasets covering bi-temporal to multi-temporal sequences, while DVL-Bench~\cite{xuan2025dynamicvl} targets long-term urban dynamics with high-resolution imagery organized into six urban understanding tasks. Despite these valuable efforts, existing benchmarks remain primarily task-oriented, capable of revealing that MLLMs perform poorly on specific long-term remote sensing tasks, yet offering limited insight into which underlying cognitive competencies are deficient. To address these gaps, we construct ChronoBench, a multi-dimensional, multi-granularity evaluation framework organized around a four-level cognitive hierarchy.

\section{ChronoBench}
To comprehensively evaluate the capabilities of MLLMs in long-term earth observation, we introduce ChronoBench, a high-resolution, long-temporal remote sensing benchmark that probes spatio-temporal intelligence across a four-level cognitive hierarchy at multiple granularities, from class-level transitions to object-level trajectory tracking. 

\begin{figure*}[tb]
    \centering
    \includegraphics[width=1\linewidth]{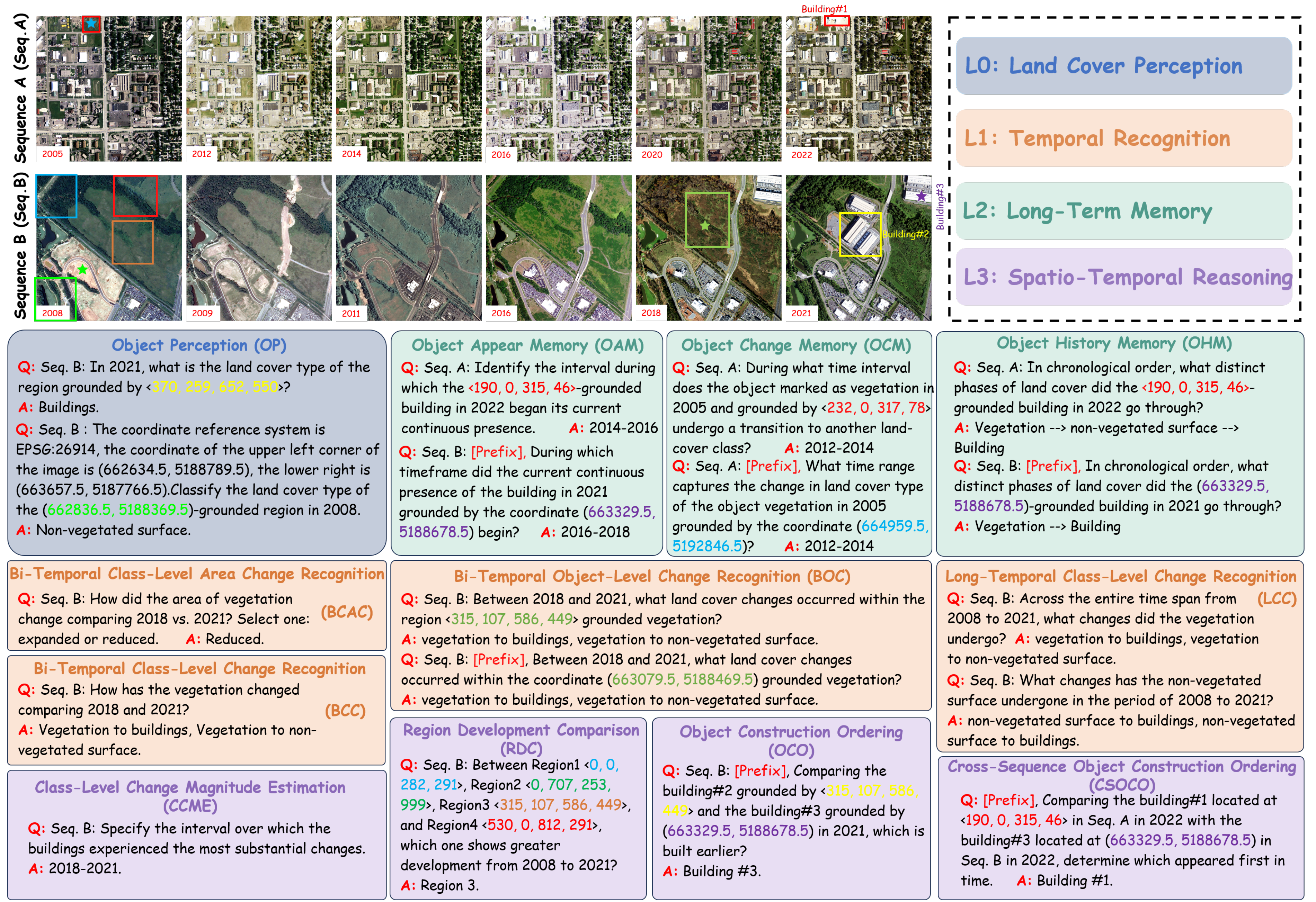}
    \caption{Task taxonomy in ChronoBench. Except for the Cross-Sequence Object Construction Ordering task, models process only one image sequence~(from either Seq. A or Seq. B) for all other tasks. Note that the labels ``Seq. A / Seq. B'' shown before task prompts are for conceptual illustration only. Additionally, the [Prefix] denotes the introduction of the coordinate reference system, along with the geographic coordinates of the top-left and bottom-right corners of the input image.}
    \label{fig:dataset}
\end{figure*}

\subsection{Task Taxonomy}\label{data: type}
We decompose long-term remote sensing understanding into a four-level cognitive hierarchy, each framed by a core question. \textbf{Level 0 Land Cover Perception}: \textit{What is here at a given time?} \textbf{Level 1 Temporal Recognition}: \textit{What changed between specified time points?} \textbf{Level 2 Long-Term Memory}: \textit{When did changes occur, and what is the complete evolution history?} \textbf{Level 3 Spatio-Temporal Reasoning}: \textit{What can be inferred by comparing across locations and time?} Note that Levels 1 and 2 differ in a key aspect: Level 1 takes time points as given inputs, whereas Level 2 requires the model to recall temporal information as outputs. Guided by this hierarchy, we design 12 sub-tasks across the four levels, as illustrated in Figure~\ref{fig:dataset}.

\textbf{Level 0: Land Cover Perception.} 
We design a basic Object Perception~(OP) task to assess whether the model can identify the land cover present at a specified location and timestamp.
To ensure robust spatial anchoring, we employ two localization modalities: normalized horizontal bounding boxes~(HBB), where coordinates are scaled within $\left[0, 999 \right]$, and geographic coordinates, where the target location is defined within the image extent specified by its upper-left and lower-right corner coordinates.

\textbf{Level 1: Temporal Recognition.} 
We evaluate this capability at both class-level and object-level granularities. At the class level, \textbf{i)} Bi-Temporal Class-Level Change Recognition~(BCC) requires comparing two specified timestamps to identify all transition types undergone by a given land cover category, and \textbf{ii)} Bi-Temporal Class-Level Area Change Recognition~(BCAC) requires determining whether its overall spatial extent has expanded or contracted. \textbf{iii)} Long-Temporal Class-Level Change Recognition~(LCC) extends this to the full observation period, aggregating all transitions a specified category has undergone across the entire sequence. At the object level, \textbf{iv)} Bi-Temporal Object-Level Change Recognition~(BOC) narrows the scope to a localized region defined by HBB Box or Geographic Coordinate, requiring identification of all land cover transitions within this area between two given time points.

\textbf{Level 2: Long-Term Memory.} 
We design three sub-tasks with increasing complexity. \textbf{i)} Object Appear Memory~(OAM) asks when a specified land cover type first appeared at a given location. \textbf{ii)} Object Change Memory~(OCM) asks when the land cover at a given location changed to a different category. \textbf{iii)} Object History Memory~(OHM) poses the most demanding challenge, requiring the model to reconstruct the complete chronological sequence of distinct land cover phases at a given location.

\textbf{Level 3: Spatio-Temporal Reasoning.} 
We design four sub-tasks that require cross-location or cross-temporal comparative reasoning. \textbf{i)} Object Construction Ordering~(OCO) presents two localized objects and requires determining which was constructed earlier. \textbf{ii)} Cross-Sequence Object Construction Ordering~(CSOCO) extends this to a cross-scene setting where the two objects reside in different geographic regions with independent image sequences. \textbf{iii)} Region Development Comparison~(RDC) specifies four regions using HBB boxes and requires identifying the one that exhibits the highest degree of urban development. \textbf{iv)} Class-level Change Magnitude Estimation~(CCME) requires the model to compare the magnitude of transitions for a specified land cover category across all consecutive time intervals and identify the period of most substantial change.

\subsection{Data Statistics and Characteristics}\label{data: statistic}
ChronoBench spans 500 distinct regions across 39 major U.S. cities, capturing the rich land-cover dynamics driven by rapid urbanization. The benchmark comprises 3,469 high-resolution images~(1024$\times$1024) and a total of 17,689 rigorously validated QA pairs, yielding an average of 35.38 QA pairs per image sequence. The QA pairs are distributed across the four levels: Land Cover Perception~(7.5\%), Temporal Recognition~(54.2\%), Long-Term Memory~(26.0\%), and Spatio-Temporal Reasoning~(12.3\%). Each QA pair uses an average of 7.26 temporal frames as visual input, mostly between 6 and 8, and up to 19. The benchmark also incorporates cross-sequence QA pairs that require the model to jointly reason over two image sequences from different geographic regions. To accommodate varying levels of task complexity, three question formats are employed: single-choice questions~(62.9\%), multiple-choice questions~(32.1\%), and ordered-sequence questions~(5.0\%) requiring chronologically faithful output. In parallel, we construct ChronoInstruct, a large-scale instruction-tuning dataset comprising 1,534 image sequences and 104,949 QA pairs, with three answer formats: multiple-choice, short-text, and free-form natural-language responses. Additional statistical analyses are provided in the supplementary material.

\begin{figure}[tb]
    \centering
    \includegraphics[width=1\linewidth]{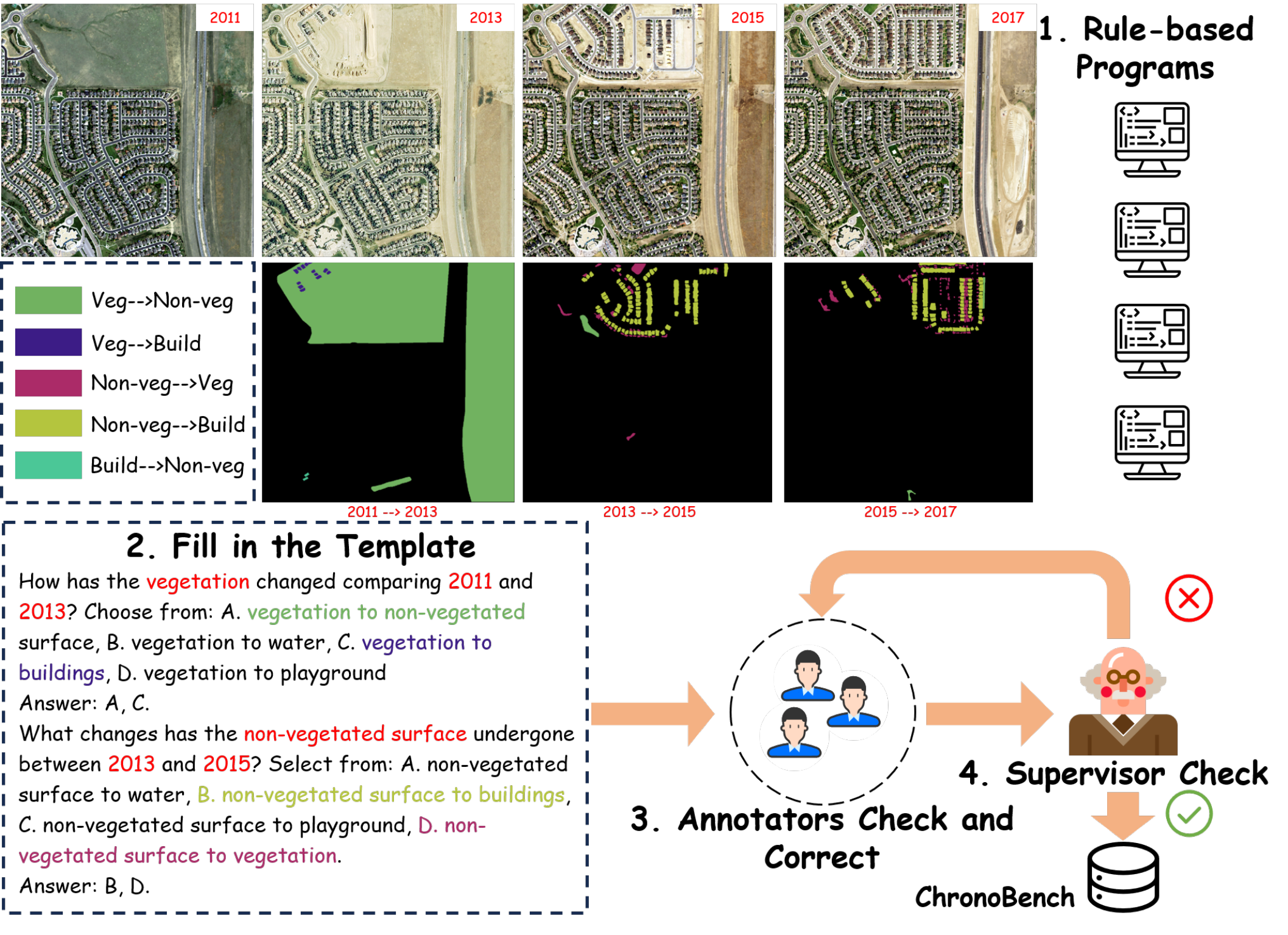}
    \caption{Overview of the data construction pipeline.}
    \label{fig:pipeline}
\end{figure}

\subsection{Data Construction Pipeline}\label{data: construction}
To ensure annotation reliability, we adopt a fully rule-based construction pipeline grounded in human-annotated semantic change masks, as is shown in Figure~\ref{fig:pipeline}. ChronoBench builds on the metadata provided by the DynamicVL team~\cite{xuan2025dynamicvl}, which includes pixel-level semantic change masks annotated at consecutive timestamps. These human-annotated masks provide a reliable basis for inferring land-cover states and transition events at arbitrary spatial locations along the temporal axis.

For each sub-task, we design deterministic rule-based programs that traverse the semantic masks to extract ground-truth answers, and pair them with 5 stylistically diverse prompt templates to ensure linguistic variety~(Figure~\ref{fig:pipeline}). All prompt templates are carefully calibrated to eliminate potential ambiguities, and the order of answer options is fully randomized to prevent positional bias.


The automatically generated QA pairs undergo a two-stage human quality control process. In the first stage, multiple independent annotators review each sample against the source images and semantic masks, flagging entries with ambiguous visual evidence, incorrect ground-truth labels, or ill-formed questions. In the second stage, a domain-expert supervisor adjudicates a random subset of flagged samples, making accept-or-reject decisions to ensure the benchmark meets a rigorous quality standard.

ChronoInstruct shares the same construction pipeline for its multiple-choice and short-text answer components. For free-form natural-language responses, we further employ Gemini-3-Flash~\cite{team2023gemini} to generate answers from the extracted ground-truth information, followed by the same two-stage human quality-control process to ensure the reliability. Additional annotation details are provided in the supplementary material.

\section{How Well Do MLLMs Understand Long-Term Remote Sensing Dynamics?}


\subsection{Evaluation Setting}
To establish a comprehensive comparison, we evaluate representative models across three distinct categories: \textbf{i)} \textbf{Commercial MLLMs}, i.e., Gemini-3-Flash~\cite{team2023gemini}, Seed-1.6-Vision~\cite{bytedance2026seed16} and GPT-5.4~\cite{openai2025gpt41}; \textbf{ii)} \textbf{Open-source General MLLMs}, comprising the InternVL-3.5 series~(4B, 8B, 14B)~\cite{wang2025internvl3}, and the Qwen3-VL series~(4B, 8B, 32B)~\cite{bai2025qwen3}; and \textbf{iii)} \textbf{Remote Sensing Domain Models} with temporal remote sensing understanding capabilities, i.e., TEOChat~\cite{irvin2024teochat}, EarthDial~\cite{soni2025earthdial}, and DVLChat~\cite{xuan2025dynamicvl}. Additionally, we recruit three domain experts to establish human-level performance, providing an upper-bound reference for evaluating model capabilities.

\begin{table*}[tb]
\centering
\caption{Quantitative evaluation results on ChronoBench. We compare human expert performance against commercial MLLMs, open-source general MLLMs, remote sensing domain models, and our GeoChrono. All values are reported in percentage (\%).}
\label{tab:main_table}
\setlength{\tabcolsep}{3pt}
\renewcommand{\arraystretch}{1.10}
\resizebox{\textwidth}{!}{
\begin{tabular}{
l
c c c!{\vrule width 1pt}
c c c c c c!{\vrule width 1pt}
c c c c c c c!{\vrule width 1pt}
c c c c c c c!{\vrule width 1pt}
c
}
\toprule
\multirow{3}{*}{\textbf{Method}}
& \multicolumn{3}{>{\columncolor{LCPColor}}c!{\vrule width 1pt}}{\textbf{Land Cover Perception}}
& \multicolumn{6}{>{\columncolor{TAColor}}c!{\vrule width 1pt}}{\textbf{Temporal Recognition}}
& \multicolumn{7}{>{\columncolor{LTMColor}}c!{\vrule width 1pt}}{\textbf{Long-Term Memory}}
& \multicolumn{7}{>{\columncolor{STRColor}}c!{\vrule width 1pt}}{\textbf{Spatio-Temporal Reasoning}}
& \multirow{3}{*}{\textbf{OA}} \\
\cmidrule(lr){2-4}
\cmidrule(lr){5-10}
\cmidrule(lr){11-17}
\cmidrule(lr){18-24}
& \multicolumn{2}{c}{\textbf{OP}}
& \multirow{2}{*}{\textbf{AVG}}
& \multirow{2}{*}{\textbf{BCAC}}
& \multirow{2}{*}{\textbf{BCC}}
& \multirow{2}{*}{\textbf{LCC}}
& \multicolumn{2}{c}{\textbf{BOC}}
& \multirow{2}{*}{\textbf{AVG}}
& \multicolumn{2}{c}{\textbf{OAM}}
& \multicolumn{2}{c}{\textbf{OCM}}
& \multicolumn{2}{c}{\textbf{OHM}}
& \multirow{2}{*}{\textbf{AVG}}
& \multirow{2}{*}{\textbf{CCME}}
& \multirow{2}{*}{\textbf{RDC}}
& \multicolumn{2}{c}{\textbf{OCO}}
& \multicolumn{2}{c}{\textbf{CSOCO}}
& \multirow{2}{*}{\textbf{AVG}}
& \\
\cmidrule(lr){2-3}
\cmidrule(lr){8-9}
\cmidrule(lr){11-12}
\cmidrule(lr){13-14}
\cmidrule(lr){15-16}
\cmidrule(lr){20-21}
\cmidrule(lr){22-23}
& \textbf{Coord} & \textbf{Box}
&
&
&
&
& \textbf{Coord} & \textbf{Box}
&
& \textbf{Coord} & \textbf{Box}
& \textbf{Coord} & \textbf{Box}
& \textbf{Coord} & \textbf{Box}
&
&
&
& \textbf{Coord} & \textbf{Box}
& \textbf{Coord} & \textbf{Box}
&
& \\
\midrule
\textbf{Human}
& 98.52 & 95.56 & 97.04
& 94.81 & 82.96 & 82.96
& 98.52 & 89.63
& 89.78
& 98.52 & 97.04
& 91.11 & 87.41
& 85.93 & 90.37
& 91.73
& 78.52 & 97.78
& 99.26 & 100.00
& 99.26 & 98.52
& 95.56 & 92.28 \\
\midrule

\rowcolor{GroupGray}
\multicolumn{25}{l}{\textbf{Commercial models}} \\
\specialrule{1pt}{0pt}{0pt}
Gemini-3-Flash\cite{team2023gemini}
& 65.06 & 65.96 & 65.51
& 82.00 & 50.53 & 52.45
& 55.85 & 56.16
& 61.38
& 56.44 & 58.44
& 46.68 & 47.75
& 25.00 & 21.36
& 47.52
& 37.79 & \bestSTR{63.64}
& \bestSTR{68.05} & 78.01
& 54.36 & 66.67
& 59.89 & 57.48 \\
GPT-5.4\cite{openai2025gpt41}
& 47.07 & 40.00 & 43.53
& 80.00 & 50.76 & 53.87
& 66.13 & 60.00
& 67.57
& 47.69 & 41.52
& 35.36 & 46.09
& 26.14 & 20.00
& 39.21
& 30.73 & 30.51
& 53.83 & 70.60
& 52.43 & 69.12
& 50.42 & 56.29 \\
Seed-1.6-Vision\cite{bytedance2026seed16}
& 50.98 & 73.23 & 62.11
& 81.91 & 46.27 & 53.73
& 47.08 & 58.47
& 63.87
& 37.08 & 47.05
& 37.22 & 41.85
& 20.00 & 18.41
& 36.86
& 31.22 & 38.98
& 56.52 & 73.29
& 55.83 & 79.41
& 53.74 & 55.48 \\
\midrule

\rowcolor{GroupGray}
\multicolumn{25}{l}{\textbf{Open-source models}} \\
\specialrule{1pt}{0pt}{0pt}
InternVL-3.5-4B\cite{wang2025internvl3}
& 45.41 & 58.65 & 52.03
& 54.83 & 32.57 & 27.95
& 24.51 & 24.20
& 38.19
& 22.03 & 21.49
& 6.09 & 10.20
& 1.36 & 0.91
& 13.33
& 22.76 & 33.33
& 54.66 & 49.28
& 50.49 & 52.45
& 42.07 & 33.25 \\
InternVL-3.5-8B\cite{wang2025internvl3}
& 33.08 & 45.56 & 39.32
& 56.29 & 28.31 & 28.49
& 35.60 & 39.35
& 43.21
& 25.11 & 24.30
& 13.91 & 16.95
& 1.60 & 3.41
& 16.93
& 30.40 & 39.55
& 51.35 & 53.21
& 57.77 & 47.55
& 45.11 & 36.32 \\
InternVL-3.5-14B\cite{wang2025internvl3}
& 40.90 & 55.94 & 48.42
& 64.15 & 33.41 & 30.66
& 36.43 & 30.08
& 45.67
& 22.39 & 21.76
& 15.76 & 19.07
& 0.45 & 0.91
& 16.45
& 21.95 & 28.25
& 52.38 & 53.83
& 48.06 & 49.51
& 41.42 & 37.76 \\
Qwen3-VL-4B\cite{bai2025qwen3}
& 35.19 & 41.35 & 38.27
& 66.94 & 31.13 & 36.77
& 23.96 & 16.73
& 42.07
& 22.76 & 18.95
& 18.94 & 22.52
& 2.95 & 4.55
& 17.54
& 21.30 & 23.16
& 47.83 & 47.83
& 54.85 & 53.92
& 39.53 & 35.10 \\
Qwen3-VL-8B\cite{bai2025qwen3}
& 39.40 & 41.96 & 40.68
& 69.21 & 29.15 & 43.15
& 31.98 & 29.26
& 47.12
& 29.74 & 28.56
& 17.48 & 20.00
& 2.95 & 0.91
& 20.52
& 23.74 & 28.81
& 54.66 & 53.00
& 55.83 & 47.06
& 42.80 & 39.19 \\
Qwen3-VL-32B\cite{bai2025qwen3}
& 42.86 & 44.66 & 43.76
& 75.42 & 44.29 & 47.90
& 45.68 & 46.16
& 57.88
& 29.92 & 30.73
& 25.03 & 27.95
& 3.86 & 4.55
& 24.06
& 28.78 & 32.77
& 51.76 & 59.83
& 59.71 & 62.25
& 47.23 & 46.73 \\
\midrule

\rowcolor{GroupGray}
\multicolumn{25}{l}{\textbf{Remote sensing domain models}} \\
\specialrule{1pt}{0pt}{0pt}
TEOChat-7B\cite{irvin2024teochat}
& 32.63 & 36.99 & 34.81
& 48.85 & 14.23 & 4.21
& 17.27 & 24.25
& 30.07
& 24.75 & 27.83
& 4.11 & 6.62
& 1.59 & 1.14
& 14.64
& 17.40 & 27.68
& 50.93 & 53.21
& 9.22 & 22.06
& 33.35 & 26.82 \\
EarthDial-4B\cite{soni2025earthdial}
& 36.99 & 40.30 & 38.65
& 54.83 & 18.04 & 4.34
& 17.72 & 24.09
& 33.08
& 26.20 & 26.20
& 17.48 & 17.75
& 0.68 & 1.36
& 18.56
& 15.61 & 22.03
& 48.44 & 46.38
& 41.75 & 52.45
& 36.25 & 30.11 \\
DVLChat-4B\cite{xuan2025dynamicvl}
& 48.57 & 46.92 & 47.74
& 85.90 & 31.81 & 37.18
& 34.21 & 30.46
& 54.48
& 29.56 & 26.93
& 24.37 & 29.93
& 2.50 & 2.05
& 22.91
& 17.24 & 25.42
& 50.72 & 51.76
& 58.74 & 55.39
& 40.59 & 44.07 \\
\midrule

\rowcolor{GroupGray}
\multicolumn{25}{l}{\textbf{Ours}} \\
\specialrule{1pt}{0pt}{0pt}
GeoChrono
& \bestLCP{83.91} & \bestLCP{93.38} & \bestLCP{88.65}
& \bestTA{93.66} & \bestTA{65.91} & \bestTA{76.66}
& \bestTA{77.83} & \bestTA{80.27}
& \bestTA{83.03}
& \bestLTM{72.71} & \bestLTM{87.04}
& \bestLTM{68.61} & \bestLTM{75.10}
& \bestLTM{31.59} & \bestLTM{32.73}
& \bestLTM{68.10}
& \bestSTR{62.60} & 61.58
& 61.49 & \bestSTR{92.55}
& \bestSTR{75.24} & \bestSTR{92.16}
& \bestSTR{72.92} & \bestOA{78.34} \\

\bottomrule
\end{tabular}
}
\end{table*}

\subsection{Benchmarking MLLMs on ChronoBench}\label{bench: results}
The quantitative results on ChronoBench, as detailed in Table~\ref{tab:main_table}, reveal a severe performance bottleneck across all evaluated models. While commercial models hold a relative lead, all three categories perform poorly on long-term memory tasks. This universal shortfall reflects not merely a limited parameter scale, but the lack of dedicated temporal modeling.

\textbf{Substantial Human-Machine Gap}: 
Human experts achieve an average OA of 92.28\%, confirming that the benchmark tasks are unambiguous and solvable. In contrast, the best evaluated commercial model, Gemini-3-Flash, reaches only 57.48\%. Open-source models peak at 46.73\%~(Qwen3-VL-32B), and RS domain models fare no better, with DVLChat at 44.07\%, while TEOChat and EarthDial struggle at 26.82\% and 30.11\%. This gap confirms that long-term RS interpretation remains an open challenge for current MLLMs.

\textbf{Long-Term Memory as the Critical Bottleneck}: Long-Term Memory stands out as the dimension with the largest human-machine gap. Human experts achieve 91.73\% on this dimension, yet even the best evaluated commercial model, Gemini-3-Flash, averages only 47.52\%. Open-source general models fare far worse~(13.33\%--24.06\%), and RS domain models perform comparably~(14.64\%--22.91\%) despite being specifically trained on temporal RS data. The gap widens further on Object History Memory~(OHM), which requires reconstructing complete evolution trajectories: open-source models nearly collapse~(0.45\%--4.55\%), RS domain models stay below 2.50\%, and even Gemini-3-Flash manages only 25\%. This universal failure reveals that current MLLMs lack the capacity to accurately memorize and recall long-term evolutionary trajectories of geographic entities.

\textbf{Inconsistent Scaling Effect in General MLLMs}: While expanding model parameters generally improves the Overall Average~(OA), e.g., Qwen3-VL scales from 35.10\%~(4B) to 46.73\%~(32B), and InternVL-3.5 from 33.25\%~(4B) to 37.76\%~(14B), this benefit is inconsistent in specific capability dimensions. Notably, within the InternVL series, scaling up does not yield uniform improvements; its performance on memory and reasoning tasks fluctuates or even degrades as parameter size increases. Furthermore, even when a positive correlation exists~(as in Qwen3-VL), the absolute performance still remains in a bottleneck. This underscores that brute-force parameter scaling is not a silver bullet, and bridging the massive gap to human cognition requires methodological innovation.
\vspace{-5pt}



\begin{figure*}[tb]
    \centering
    \includegraphics[width=1\linewidth]{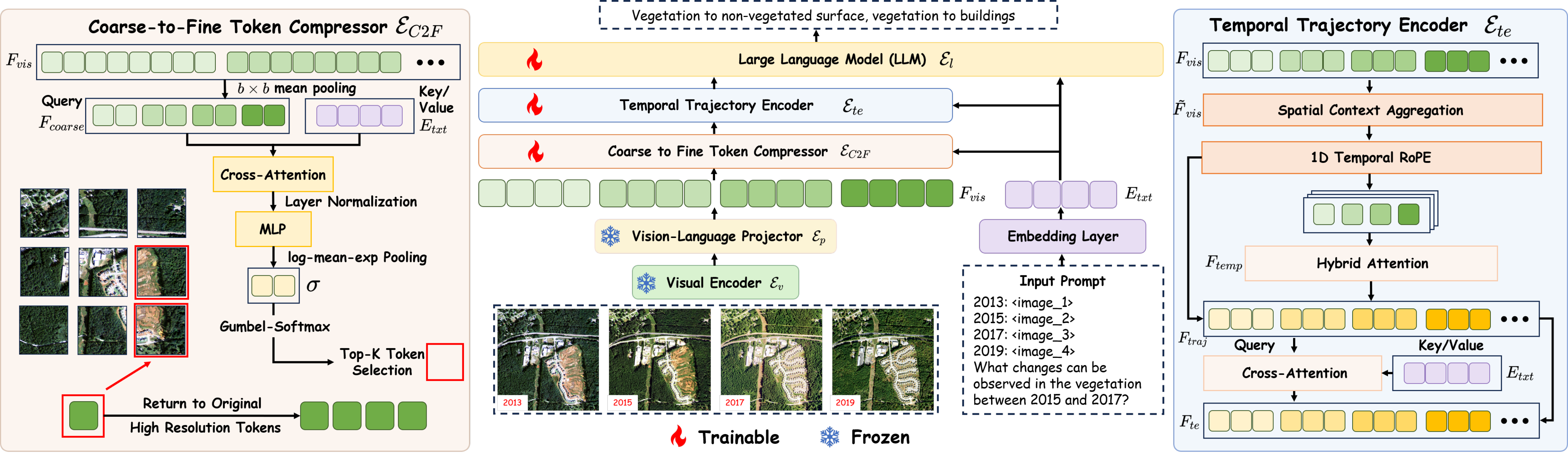}
    \caption{Overview of the GeoChrono framework. The middle panel illustrates the overall architecture. The right panel details the Temporal Trajectory Encoder~(TempEnc), which decouples the spatio-temporal feature volume into per-location temporal trajectories via hybrid bidirectional-causal attention and text-guided semantic focusing. The left panel details the Coarse-to-Fine Token Compressor~(C2FComp), which leverages text-guided saliency scoring to selectively preserve fine-grained tokens for task-relevant regions while compressing the background.}
    \label{fig: framework}
\end{figure*}

\section{GeoChrono}
As shown in Section~\ref{bench: results}, existing MLLMs struggle with long-term remote sensing dynamics, especially in long-term memory. To address this, we introduce GeoChrono, an MLLM tailored for long-term remote sensing interpretation. Given an image sequence $I_s = \{I_1, I_2, \ldots, I_T\}$ spanning $T$ frames and a text prompt $Q$, GeoChrono generates a response $A$ through spatio-temporal reasoning.

GeoChrono comprises four core components: \textbf{i)}~a Visual Encoder $\mathcal{E}_{v}$ that extracts per-frame visual features, \textbf{ii)}~a Vision-Language Projector $\mathcal{E}_{p}$ that maps visual representations into the language embedding space, \textbf{iii)}~a \textbf{Temp}oral Trajectory \textbf{Enc}oder~(TempEnc) $\mathcal{E}_{te}$ that models per-location temporal evolution to strengthen the perception of land-cover change history, and \textbf{iv)}~a Large Language Model~(LLM) $\mathcal{E}_{l}$ that performs multimodal reasoning and generates the textual response $A$ as Figure~\ref{fig: framework} shows.
Specifically, TempEnc takes the projected visual features $F_{vis}\in \mathbb{R}^{T \times S \times D}$~($T$: temporal frames, $S$: spatial tokens per frame, $D$: embedding dimension) with text instruction embeddings $E_{txt} \in \mathbb{R}^{L\times D}$, and outputs temporally enhanced representations $F_{te}$ that are fed into the LLM. Beyond the core architecture, we further introduce a \textbf{C}oarse-to-\textbf{F}ine Token \textbf{Comp}ressor~(C2FComp) $\mathcal{E}_{C2F}$ that alleviates the computational burden of long high-resolution sequences by adaptively compressing static backgrounds while preserving task-relevant dynamic features. Sections~\ref{sec:ttc} and~\ref{sec:c2f} detail TempEnc and C2FComp, respectively.
\vspace{-6pt}

\subsection{Temporal Trajectory Encoder}\label{sec:ttc}
Unlike natural videos, temporal remote sensing sequences exhibit a distinctive geostationary prior: the sensor's perspective varies negligibly, so each spatial patch across frames corresponds to the same geographic location. This allows the semantic evolution at any location to be modeled as an independent 1D temporal trajectory tracking land-cover change at a fixed spatial point. TempEnc explicitly exploits this prior by decoupling the entangled spatio-temporal feature volume into per-location trajectories for dedicated temporal modeling. It comprises three stages: \textbf{Spatial Context Aggregation}, which enriches each token with local spatial context; \textbf{Hybrid Temporal Attention}, which models per-location dynamics via hybrid attention; and \textbf{Semantic Focusing}, which uses text-guided cross-attention to suppress task-irrelevant temporal information.

\textbf{Spatial Context Aggregation.} Constructing trajectories from isolated patches neglects the regional continuity inherent in remote sensing land cover. To inject local spatial context before trajectory formation, we apply a lightweight $3 \times 3$ depthwise convolution over the 2D-reshaped features as a pre-norm residual block:
\begin{align}
\tilde{F}_{vis}=F_{vis}+\text{Flatten}\Big(\text{Conv}_{3\times3}\big(\text{Reshape}(\text{LN}(F_{vis}))\big)\Big)\,,
\end{align}
where $\text{LN}(\cdot)$ denotes Layer Normalization and $\text{Reshape}$ transposes the spatial tokens into a 2D grid.

\textbf{Hybrid Temporal Attention.} From the enriched features $\tilde{F}_{vis}\in \mathbb{R}^{T \times S \times D}$, we extract per-location temporal trajectories by grouping tokens sharing the same spatial index across all frames:
\begin{align}
z_s = \big[\tilde{F}_{vis}^{1,s};\; \tilde{F}_{vis}^{2,s};\; \ldots;\; \tilde{F}_{vis}^{T,s}\big] \in \mathbb{R}^{T \times D}\,,
\end{align}
where each $z_s$ encapsulates the complete temporal evolution at a single geographic location, and $\big[;\big]$ denotes concatenation.

We first inject 1D Rotary Position Embeddings~(RoPE)~\cite{su2024roformer} along the temporal axis of $z_s$ to encode relative temporal distances. We then model the temporal trajectories with dual-stream hybrid attention, where the $H$ attention heads are partitioned into two equal streams: the first $H/2$ heads apply \textit{bidirectional} self-attention $\mathcal{G}_{bi}$ over the full temporal sequence to capture global change patterns and long-range correlations, yielding $A_{bi}=\mathcal{G}_{bi}(z_s) \in \mathbb{R}^{T \times D/2}$, while the remaining $H/2$ heads apply \textit{causal} self-attention $\mathcal{G}_{ca}$ to preserve chronological progression, yielding $A_{ca}=\mathcal{G}_{ca}(z_s) \in \mathbb{R}^{T \times D/2}$. The two streams are concatenated and projected to produce the hybrid attention output $\mathcal{H}(\mathbf{z}_s)$ for each layer:
\begin{align} 
\mathcal{H}(\mathbf{z}_s) = \big[A_{bi};\; A_{ca}\big]\, W_O\,,
\end{align}
where $W_O \in \mathbb{R}^{D \times D}$ is the output projection matrix. After $L$ layers, the output trajectories are reassembled into $F_{temp} \in \mathbb{R}^{T \times S \times D}$ and combined via a residual connection:
\begin{align}
F_{traj} = F_{temp} + \tilde{F}_{vis}\,.
\end{align}

\textbf{Semantic Focusing.} Remote sensing long temporal sequences inevitably carry substantial task-irrelevant information. We employ cross-attention with $F_{traj}$ as Query and $E_{txt}$ as Key/Value, allowing the text instruction to act as a semantic filter that selectively amplifies task-relevant temporal cues:
\begin{equation}
\begin{aligned}
\text{CrossAttn}(F_{traj}, E_{txt})
= \mathrm{softmax}\!\left(\frac{(F_{traj} W_Q)(E_{txt} W_K)^\top}{\sqrt{d_k}}\right) (E_{txt} W_V)\,,
\end{aligned}
\end{equation}
where $W_Q$, $W_K$, and $W_V$ are learnable projection matrices. The final output of TempEnc $\mathcal{E}_{te}$ is:
\begin{align}
    F_{te} = F_{traj} + \text{CrossAttn}(F_{traj}, E_{txt})\,.
\end{align}

\subsection{Coarse-to-Fine Token Compressor}\label{sec:c2f}

High-resolution long-temporal remote sensing sequences yield a total of $T \times S$ visual tokens, which scale linearly with the number of frames, imposing a heavy computational burden on the LLM. Naive uniform compression~(e.g., global average pooling) would indiscriminately discard fine-grained details crucial for temporal understanding. We observe a key physical prior in remote sensing, \textit{change sparsity}: the land-cover transitions concentrate in a small fraction of the scene, while the majority remains static. This motivates C2FComp~($\mathcal{E}_{C2F}$), a selective compression module that retains full-resolution tokens for task-relevant dynamic regions while compressing the static background into compact coarse representations, preserving fine-grained detail for task-relevant dynamics while substantially reducing the overall token budget. Concretely, C2FComp operates through three cascaded stages, i.e., \textbf{Spatial Block Partitioning}, \textbf{Prompt-Aware Scoring}, and \textbf{Token Mixing}.

\textbf{Spatial Block Partitioning.} 
Given the fine-grained features $F_{vis} \in \mathbb{R}^{T \times S \times D}$ with spatial resolution $S = H \times W$, we group each set of $b \times b$ spatially adjacent feature vectors into a block, resulting in $S_c = \frac{H}{b} \times \frac{W}{b}$ blocks. Each block is represented by mean pooling over its constituent tokens:
\begin{align} 
F_{coarse}^{t,j} = \frac{1}{b^2} \sum_{(i,k) \in \mathcal{B}_j} F_{vis}^{t,(i,k)}\,, 
\end{align}
where $F_{coarse}^{t,j}$ is the feature for $j$-th  block $\mathcal{B}_j$  of the $t$-th frame.
This produces a block-level coarse feature map $F_{coarse} \in \mathbb{R}^{T \times S_c \times D}$.

\textbf{Prompt-Aware Scoring.} A lightweight saliency module scores each block's relevance to the text instruction embeddings. The coarse features are passed through a cross-attention layer (visual tokens $F_{coarse}$ as Query, $E_{txt}$ as Key/Value), followed by a LayerNorm-MLP scoring head that produces saliency logits $l$. These are then aggregated into a unified per-position signal via log-mean-exp pooling across the temporal axis:
\begin{align}
\sigma_j=\log(\!\frac{1}{T}\sum_{t=1}^{T} \exp(l^{t,j}))\,, 
\end{align}
yielding the selection signal $\sigma \in \mathbb{R}^{S_c}$.


\textbf{Token Mixing.} Given the saliency signal $\sigma$, we select the top-$K$ blocks with the highest scores, where $K$ is a hyperparameter that controls the overall compression rate. The selection is implemented via Gumbel-Softmax relaxation~\cite{jang2016categorical} with straight-through estimation for differentiable training, and reduces to deterministic hard top-$K$ at inference. The resulting binary mask $\mathbf{m} \in \{0,1\}^{S_c}$ is shared across all $T$ frames to ensure spatially consistent compression. For the $j$-th  block of the $t$-th frame, the mixed representation is:
\begin{align}
F_{mixed}^{t,j} = \begin{cases} F_{fine}^{t,j} \in \mathbb{R}^{b^2 \times D}, & \text{if } \mathbf{m}_j = 1 \\ F_{coarse}^{t,j} \in \mathbb{R}^{1 \times D}, & \text{if } \mathbf{m}_j = 0 \end{cases}
\end{align}
where $F_{fine}^{t,j}$ denotes the $b^2$ fine-grained tokens within the $j$-th block. Concatenating all blocks yields $F_{mixed} \in \mathbb{R}^{T \times N_{mixed} \times D}$ with $N_{mixed} = K \cdot b^2 + (S_c - K) \ll S$, which is then forwarded to TempEnc for temporal trajectory modeling.

\section{Experiments}\label{sec:experiments}

We first describe implementation settings~(Section~\ref{sec:impl}), then compare GeoChrono against existing MLLMs and evaluate its generalizability on external benchmarks~(Section~\ref{sec:main_exp}), followed by an efficiency analysis of C2FComp~(Section~\ref{sec: c2f}) and ablation studies~(Section~\ref{sec: ablation}). Detailed training configurations, qualitative results, and case studies are provided in the supplementary material.

\subsection{Implementation Settings}\label{sec:impl}

GeoChrono is built upon Qwen3-VL-4B-Instruct~\cite{bai2025qwen3}. During training, the Vision Encoder and Vision-Language Projector are frozen; TempEnc and C2FComp are randomly initialized and fully fine-tuned, while the LLM backbone is tuned via LoRA~\cite{hu2022lora}. Both TempEnc and C2FComp adopt a single-layer attention architecture. The model is trained for one epoch on ChronoInstruct using 4 NVIDIA H100 80GB GPUs. Full hyperparameter details are provided in the supplementary material.

\begin{table}[tb]
\centering
\caption{Zero-shot evaluation on DVL-Bench~\cite{xuan2025dynamicvl} and CDVQA~\cite{yuan2022change}. All values are reported in percentage (\%).}
\label{tab:zero_shot}
\small
\setlength{\tabcolsep}{4.5pt}
\renewcommand{\arraystretch}{1.08}
\begin{tabular}{lccc!{\vrule width 0.8pt}c}
\toprule
\multirow{2}{*}{\textbf{Method}} 
& \multicolumn{3}{c!{\vrule width 0.8pt}}{\textbf{DVL-Bench}} 
& \multirow{2}{*}{\textbf{CDVQA}} \\
\cmidrule(lr){2-4}
& \textbf{BCA-Single} & \textbf{BCA-Multi} & \textbf{CSE-Single} & \\
\midrule
TEOChat\cite{irvin2024teochat}   & 35.1 & 8.7  & 17.0 & 50.0 \\
EarthDial\cite{soni2025earthdial} & 62.2 & 20.3 & 30.9 & \underline{52.1} \\
DVLChat\cite{xuan2025dynamicvl}   & \underline{64.9} & \underline{21.3} & \underline{31.3} & 43.7 \\
\midrule
GeoChrono & \textbf{72.9} & \textbf{42.1} & \textbf{35.7} & \textbf{59.2} \\
\bottomrule
\end{tabular}
\end{table}

\subsection{Evaluation of GeoChrono}\label{sec:main_exp}

As shown in Table~\ref{tab:main_table}, GeoChrono achieves 78.34\% overall accuracy, surpassing leading commercial MLLMs by over 20\% and all open-source and RS domain models by an even wider margin. GeoChrono leads across all four dimensions, with the most pronounced gain in Long-Term Memory~(68.10\%), substantially narrowing the gap with human experts~(91.73\%). This improvement stems from TempEnc's hybrid temporal trajectory attention: the causal stream preserves chronological progression along each spatial trajectory for faithful recall of historical states, while the bidirectional stream captures global temporal contrasts that disambiguate similar land-cover phases across distant time steps. Consistent advantages in Temporal Recognition~(83.03\%) and Spatio-Temporal Reasoning~(72.92\%) further validate the effectiveness of per-location trajectory decoupling with text-guided semantic focusing.

We further evaluate generalizability on two external benchmarks~(Table~\ref{tab:zero_shot}). On DVL-Bench~\cite{xuan2025dynamicvl}, GeoChrono achieves 72.9\%, 42.1\%, and 35.7\% on the three sub-tasks under a zero-shot setting, consistently surpassing DVLChat~(64.9\%, 21.3\%, 31.3\%) which was trained on DVL-Instruct. On CDVQA~\cite{yuan2022change}, GeoChrono attains 59.2\% zero-shot accuracy, outperforming all RS domain baselines. These results confirm that GeoChrono's temporal modeling capabilities generalize effectively to unseen tasks and distributions.

\begin{table}[tb]
\centering
\caption{Impact of different coarse-to-fine selection ratios on efficiency and performance.}
\label{tab:c2f_ratio}
\small
\setlength{\tabcolsep}{3.8pt}
\renewcommand{\arraystretch}{1.08}
\resizebox{\columnwidth}{!}{
\begin{tabular}{lcccc}
\toprule
\textbf{Selection} & \textbf{FLOPS} & \textbf{Visual Token} & \textbf{OA} & \textbf{Performance} \\
\textbf{Ratio} &  & \textbf{Length} &  & \textbf{Ratio} \\
\midrule
w/o C2FComp     & 100.00\%  & 100.00\%  & 78.34 & 100.00\% \\
768 $\times$ 768 & -48.75\% & -43.75\%  & 73.95 & 94.40\%  \\
\midrule
1/2             & -39.22\%  & -37.50\%  & 74.58 & 95.20\%  \\
1/4             & -56.53\%  & -56.25\%  & 74.11 & 94.60\%  \\
1/16            & -68.49\%  & -70.31\%  & 72.60 & 92.67\%  \\
1/64            & -71.34\%  & -73.83\%  & 71.56 & 91.35\%  \\
\bottomrule
\end{tabular}
}
\end{table}

\begin{table}[tb]
\centering
\caption{Ablation study. DFT denotes direct fine-tuning. 
TAH, TAB, and TAC denote hybrid, bidirectional-only, and causal-only temporal attention, respectively. SAG and SFC denote spatial context aggregation and semantic focusing, respectively. LCP, TR, LTM, and STR denote land cover perception, temporal recognition, long-term memory, and spatio-temporal reasoning, respectively.}
\label{tab:ablation_temporal}
\small
\setlength{\tabcolsep}{3.5pt}
\renewcommand{\arraystretch}{1.08}
\resizebox{\columnwidth}{!}{
\begin{tabular}{lccccc}
\toprule
\textbf{Method} & \textbf{LCP} & \textbf{TR} & \textbf{LTM} & \textbf{STR} & \textbf{OA} \\
\midrule
DFT                       & 82.03 & 79.08 & 59.23 & 65.82 & 72.52 \\
DFT + TAH                 & 85.49 & 81.49 & 62.77 & 70.20 & 75.54 \\
DFT + TAH + SAG           & 85.34 & 81.62 & 63.00 & 71.59 & 75.83 \\
\midrule
DFT + TAB + SAG + SFC     & 86.02 & \underline{82.55} & 64.77 & \textbf{73.80} & 77.12 \\
DFT + TAC + SAG + SFC     & \underline{86.39} & 82.50 & \underline{65.62} & 72.60 & \underline{77.19} \\
DFT + TAH + SAG + SFC \textbf{(Full)}
                          & \textbf{88.65} & \textbf{83.03} & \textbf{68.10} & \underline{72.92} & \textbf{78.34} \\
\bottomrule
\end{tabular}
}
\end{table}

\subsection{Effectiveness of C2FComp}\label{sec: c2f}
Table~\ref{tab:c2f_ratio} studies C2FComp under varying selection ratios, with the full model as the upper bound and 768$\times$768 uniform resizing as a naive baseline. C2FComp demonstrates strong robustness across all compression levels, consistently retaining over 91\% of the full model's performance even at the most aggressive 1/64 ratio, which reduces the visual tokens by 73.83\%. The 1/4 ratio further provides over 56\% token reduction while preserving 94.60\% performance, surpassing the 768$\times$768 resizing baseline in both compression rate~(56.25\% vs.\ 43.75\%) and performance retention~(94.60\% vs.\ 94.40\%), confirming that instruction-guided selective compression is more effective than uniform downsampling.

\subsection{Ablation Studies}\label{sec: ablation}
Table~\ref{tab:ablation_temporal} progressively ablates GeoChrono's key designs. Introducing hybrid temporal attention over the DFT baseline yields 3.02\% OA gain, confirming the necessity of explicit per-location temporal modeling, while spatial context aggregation provides a further modest improvement. Among the three attention variants~(Bidirectional-only, Causal-only, and Hybrid) equipped with text-guided semantic focusing, bidirectional-only excels in Spatio-Temporal Reasoning~(73.80\%) via its global temporal view, while causal-only leads in Long-Term Memory~(65.62\%) by preserving chronological order; their hybrid combination in GeoChrono surpasses both with overall accuracy of 78.34\%, demonstrating that the two streams capture complementary temporal patterns. The consistent 2.51\% gain over the variant without semantic focusing further validates its effectiveness in filtering task-irrelevant temporal information.

\section{Conclusion}
This work rethinks long-term temporal understanding in remote sensing by proposing ChronoBench, a benchmark that decomposes this challenge into four progressive cognitive levels. Evaluations on ChronoBench identify Long-Term Memory as the most critical bottleneck of current MLLMs. We further construct ChronoInstruct, a 104K-sample instruction-tuning dataset spanning all competency levels, and propose GeoChrono, which exploits the physical priors unique to remote sensing through a Temporal Trajectory Encoder for per-location evolution modeling and a Coarse-to-Fine Token Compressor for selective visual token compression. GeoChrono achieves an overall accuracy of 78.34\% on ChronoBench, surpassing the leading commercial MLLMs by more than 20\%, and C2FComp reduces visual tokens by more than 56\% while retaining 94.6\% of full-model performance. We hope this work inspires the community to further explore this challenging direction.

\begin{acks}
This work has been funded by the National Natural Science Foundation of China under Grant 62301063.
\end{acks}

\bibliographystyle{ACM-Reference-Format}
\bibliography{sample-base}

\clearpage
\onecolumn
\appendix
\suppressfloats[t]

\begin{center}
    {\LARGE\bfseries Appendix}
\end{center}
\vspace{6pt}

{\large\bfseries Table of Contents}
\vspace{2pt}
\hrule
\vspace{4pt}
\startcontents[appendix]
\printcontents[appendix]{l}{1}{\setcounter{tocdepth}{2}}
\vspace{4pt}
\hrule
\vspace{12pt}

\section{Data Construction Details}\label{sec:data_construction}

\begin{figure}[!htb]
    \centering
    \includegraphics[width=1\linewidth]{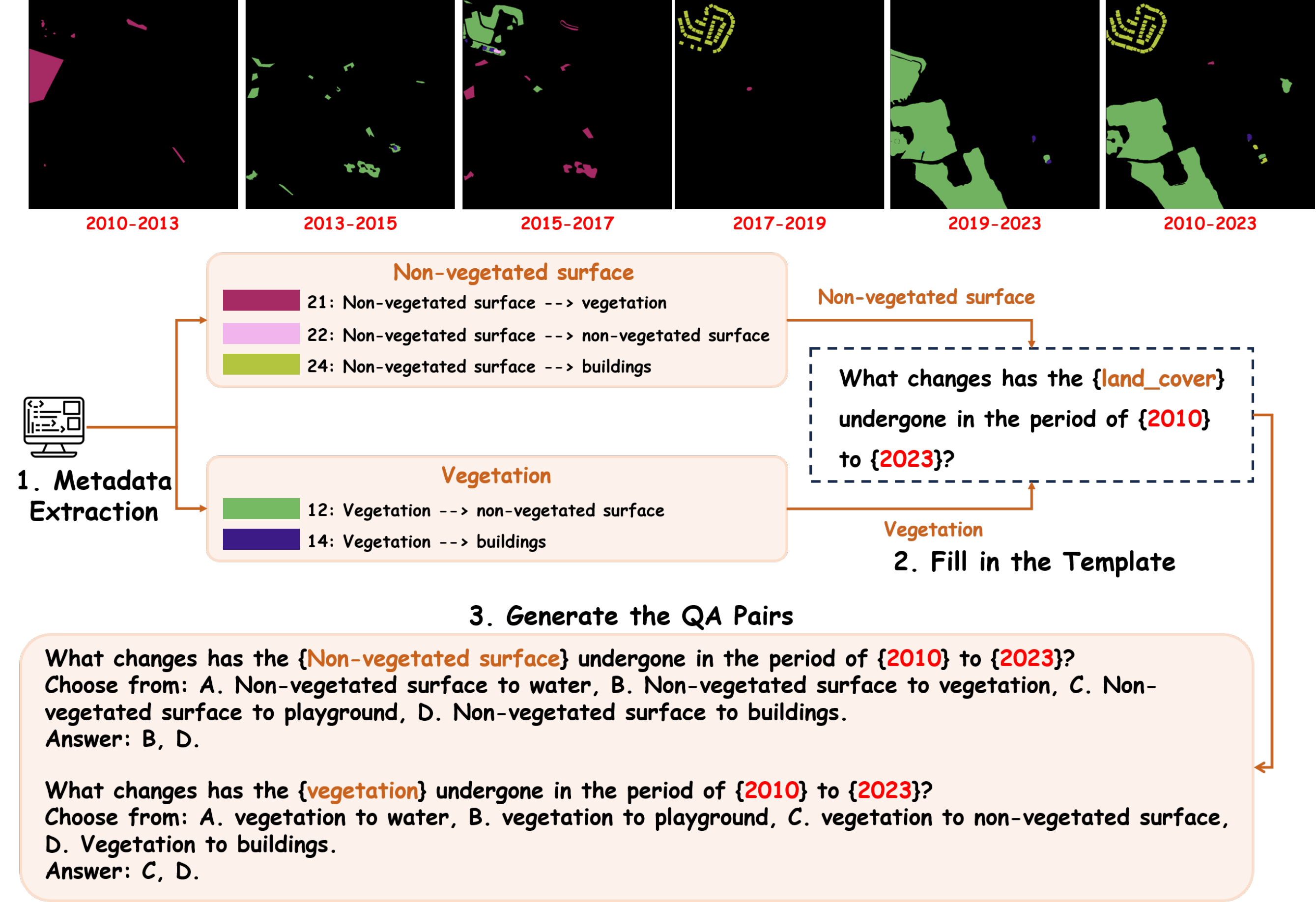}
    \caption{Illustration of the data construction pipeline, using Long-Temporal Class-Level Change Recognition~(LCC) as an example.}
    \label{fig:construction_example}
\end{figure}

\begin{figure}[!htb]
    \centering
    \includegraphics[width=1\linewidth]{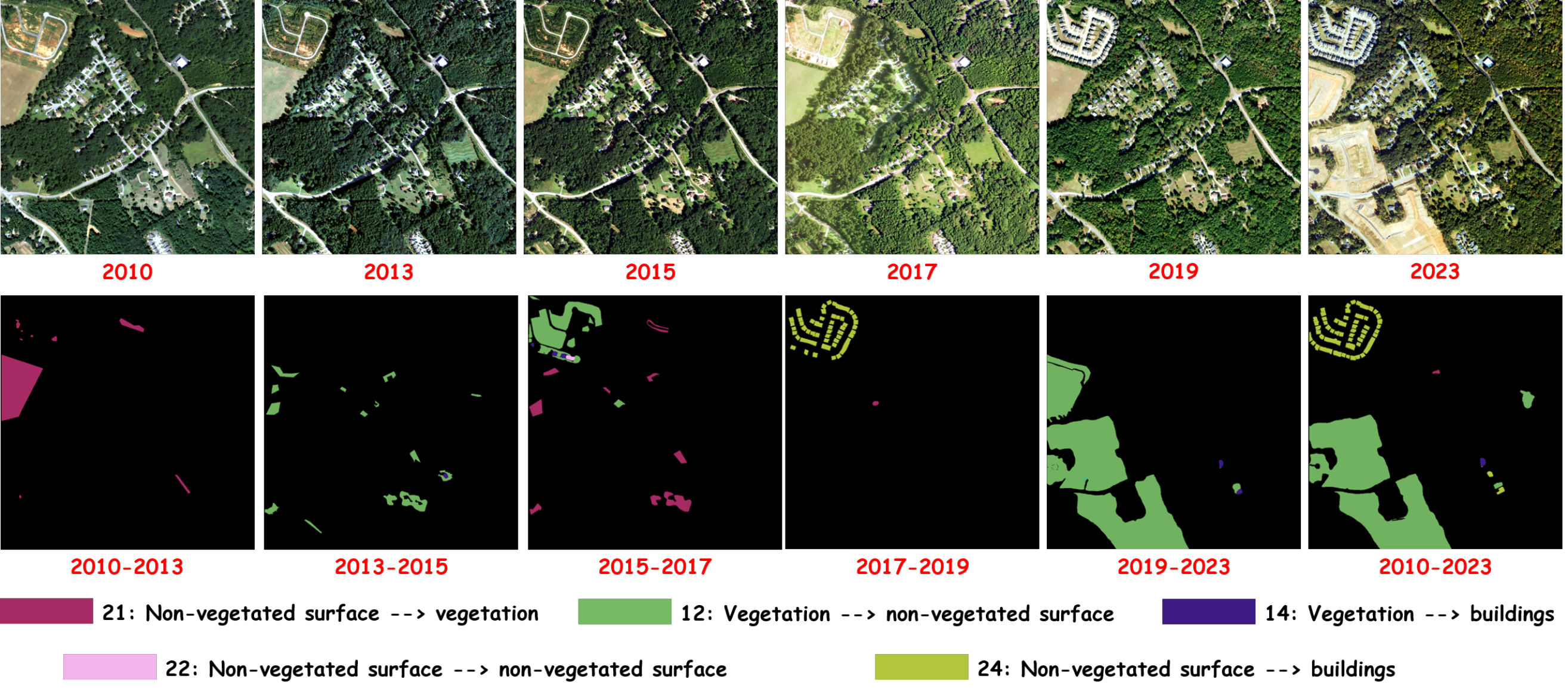}
    \caption{Illustration of the change semantic mask encoding. Each valid pixel stores a two-digit value encoding the source-to-destination land-cover transition. Background pixels (no change) are set to zero.}
    \label{fig:mask_encoding}
\end{figure}

The construction of ChronoBench and ChronoInstruct relies on deterministic rule-based programs that enable reproducible, large-scale data generation. The pipeline operates in two stages: for each geographic tile, we first execute rule-based programs to systematically analyze the human-annotated change semantic masks, extracting structured land-cover transition evidence forming the metadata; the extracted information is then populated into pre-designed prompt templates tailored to each of the 12 sub-tasks to produce the final QA pairs. 
Figure~\ref{fig:construction_example} provides a concrete end-to-end illustration of this pipeline using the LCC task as an example.
In the following, we first formalize the rule-based metadata extraction process, covering the mask encoding scheme, notation, and extraction algorithms~(Section~\ref{sec:rule_programs}); then present the prompt templates for all 12 sub-tasks~(Section~\ref{sec:templates}); and finally describe the natural-language response generation process for ChronoInstruct~(Section~\ref{sec:nl_response}).

\subsection{Rule-Based Metadata Extraction}\label{sec:rule_programs}

\subsubsection{Change Semantic Mask Encoding}\label{sec:mask_enc}

All benchmark samples are derived from semantic change masks that record pixel-level land-cover transitions between consecutive timestamps and the total transition between the first and last timestamp. As illustrated in Figure~\ref{fig:mask_encoding}, each valid pixel in a change mask stores a two-digit integer value $v$, where the tens digit encodes the \textit{source} (i.e., pre-change) land-cover class and the units digit encodes the \textit{destination} (i.e., post-change) class:
\begin{equation}\label{eq:mask_decode}
c_s = \lfloor v / 10 \rfloor, \quad c_d = v \bmod 10,
\end{equation}
where $c_s$ and $c_d$ denote the source and destination class indices, respectively. The five predefined land-cover classes and their corresponding indices are listed in Table~\ref{tab:class_mapping}. For instance, a pixel value of $14$ encodes the transition \textit{vegetation}~$\to$~\textit{buildings}, while $42$ encodes \textit{buildings}~$\to$~\textit{non-vegetated surface}. Background pixels (i.e., no change) are set to zero. During construction, only valid transition pixels are retained, satisfying: \textbf{i)}~$v$ is a two-digit code, \textbf{ii)}~$c_s \neq c_d$, and \textbf{iii)}~$c_s, c_d \in \{1,2,3,4,5\}$.

\begin{table}[t]
\centering
\caption{Land-cover class index mapping used in the change semantic masks.}
\label{tab:class_mapping}
\begin{tabular}{cl}
\toprule
Index & Land-Cover Class \\
\midrule
1 & Vegetation \\
2 & Non-vegetated Surface \\
3 & Water \\
4 & Buildings \\
5 & Playground \\
\bottomrule
\end{tabular}
\end{table}

\subsubsection{Notation and Definitions}\label{sec:notation}
For brevity, we adopt the following task abbreviations throughout this section: Object Perception~(OP), Bi-Temporal Class-Level Change Recognition~(BCC) Bi-Temporal Class-Level Area Change Recognition~(BCAC), Long-Temporal Class-Level Change Recognition~(LCC), Bi-Temporal Object-Level Change Recognition~(BOC), Object Appear Memory~(OAM), Object Change Memory~(OCM), Object History Memory~(OHM), Class-Level Change Magnitude Estimation~(CCME), Region Development Comparison~(RDC), Object Construction Ordering~(OCO), and Cross-Sequence Object Construction Ordering~(CSOCO).

The key notation used in the pseudocode is defined below:
\begin{itemize}
    \item $\mathcal{D}$: the set of all geographic tiles.
    \item $\mathcal{M}_d$: the set of all change masks associated with tile $d$.
    \item $M_{\mathrm{total}}$: the \textit{total mask}, i.e., the mask spanning the longest temporal interval, representing the cumulative land-cover change from the first to the last timestamp.
    \item $\mathcal{M}_{\mathrm{inter}} = \{M^{(1)}, M^{(2)}, \ldots\}$: the \textit{intermediate masks}, each covering a consecutive timestamp pair $[t_0^{(i)}, t_1^{(i)}]$, sorted in chronological order. These capture fine-grained, interval-specific transitions.
    \item $v_p$: the pixel value at spatial position $p$ in a change mask.
    \item $c_s(v),\, c_d(v)$: the source and destination class indices decoded from value $v$ via Eq.~\eqref{eq:mask_decode}.
    \item $\mathrm{CC}(\cdot)$: connected component extraction on a binary mask, returning a set of spatially contiguous regions.
    \item $\mathrm{Area}(R)$: the pixel area of a connected region $R$.
    \item $\mathrm{FR}(R) = \mathrm{Area}(R)\, /\, \mathrm{Area}(\mathrm{BBox}(R))$: the bounding-box fill ratio of region $R$.
    \item $\theta_a(c)$: the class-dependent minimum area threshold for retaining a candidate region, set larger for spatially extensive classes~(vegetation, non-vegetated surface) and smaller for compact classes~(water, buildings, playground).
    \item $\theta_f$: the minimum bounding-box fill-ratio threshold for region filtering.
    \item $\theta_t(c_s, c_d)$: the class-pair-dependent minimum pixel count for retaining a class-level transition.
\end{itemize}

\subsubsection{QA Metadata Extraction Algorithms}\label{sec:algorithms}

\begin{algorithm}[!b]
\caption{Shared Preprocessing}\label{alg:preprocessing}
\begin{algorithmic}[1]
\Require Tile dataset $\mathcal{D}$
\Ensure Preprocessed masks $\{(M_{\mathrm{total}},\;\mathcal{M}_{\mathrm{inter}},\;\mathcal{T})\}$ for each tile
\For{each tile $d \in \mathcal{D}$}
    \State Collect all change masks $\mathcal{M}_d = \{M_1, \ldots, M_K\}$
    \For{each $M_k \in \mathcal{M}_d$}
        \State Parse temporal interval $[t_0^{(k)}, t_1^{(k)}]$ from file name
    \EndFor
    \State $M_{\mathrm{total}} \gets \text{argmax}_{M_k} \big(t_1^{(k)} - t_0^{(k)}\big)$
    \State $\mathcal{M}_{\mathrm{inter}} \gets \mathcal{M}_d \setminus \{M_{\mathrm{total}}\}$, sorted by $t_0$
    \State $\mathcal{T} \gets$ sorted unique timestamps from all intervals
\EndFor
\end{algorithmic}
\end{algorithm}

The 12 sub-tasks share a common preprocessing stage~(Algorithm~\ref{alg:preprocessing}) and are grouped into two families based on their core extraction operation: \textbf{i)}~four \textit{class-level tasks}~(BCC, BCAC, LCC, CCME) that analyze transition pixel counts without requiring spatial connectivity, formalized in Algorithm~\ref{alg:class_level}; and \textbf{ii)}~eight \textit{object-level tasks}~(OP, BOC, OAM, OCM, OHM, OCO, RDC, CSOCO) that rely on connected-component extraction to identify discrete spatial regions, formalized in Algorithm~\ref{alg:object_level}. An important design dimension is the \textit{temporal source}: tasks assessing long-term cumulative change~(LCC, OAM, OHM, OCO, RDC) operate on $M_{\mathrm{total}}$, while tasks targeting interval-specific transitions~(BCC, BCAC, CCME, OP, BOC, OCM) operate on $\mathcal{M}_{\mathrm{inter}}$.

\begin{algorithm}[!h]
\caption{Class-Level Task Construction}\label{alg:class_level}
\begin{algorithmic}[1]
\Require Preprocessed tile, task type $\tau \in \{\text{BCC, BCAC, LCC, CCME}\}$
\Ensure Class-level QA metadata $\mathcal{R}$
\Statex \textbf{Phase 1: Transition Pixel Counting}
\If{$\tau = \text{LCC}$}
    \State $\hat{\mathcal{M}} \gets \{\text{all } \mathcal{M}_{\mathrm{inter}} \text{ merged}\}$ \Comment{aggregate over full sequence}
\Else
    \State $\hat{\mathcal{M}} \gets \mathcal{M}_{\mathrm{inter}}$ \Comment{per-interval processing}
\EndIf
\For{each mask $M \in \hat{\mathcal{M}}$}
    \For{each valid pixel $p$ in $M$ with $c_s \neq c_d$}
        \State $\Gamma[M][\,c_s(v_p) \!\to\! c_d(v_p)\,] \mathrel{+}= 1$
    \EndFor
\EndFor
\Statex \textbf{Phase 2: Task-Specific Extraction} \Comment{applied per $M$ from Phase 1}
\If{$\tau \in \{\text{BCC},\, \text{LCC}\}$}
    \For{each $(c_s \!\to\! c_d)$ with $\Gamma[M][c_s \!\to\! c_d] < \theta_t(c_s, c_d)$}
        \State Mark both $c_s$ and $c_d$ as invalid
    \EndFor
    \State Remove all transitions involving any invalid class
    \For{each valid source class $c_s$}
        \State $\mathcal{R}.\text{add}\big(c_s,\;\{c_d \mid c_s \!\to\! c_d \text{ valid}\},\;[t_0^M, t_1^M]\big)$
    \EndFor
\ElsIf{$\tau = \text{BCAC}$}
    \For{each class $c$, each mask $M$}
        \State $\Delta_c \!\gets\! \sum_{k} \Gamma[M][k \!\to\! c] - \sum_{k} \Gamma[M][c \!\to\! k]$ \Comment{net area change}
        \If{$|\Delta_c| > \theta_s(c)$}
            \State $\mathcal{R}.\text{add}(c,\, M,\, \Delta_c \!>\! 0\ ?\ \text{``expanded''}:\text{``reduced''})$
        \EndIf
    \EndFor
\ElsIf{$\tau = \text{CCME}$}
    \For{each class $c$}
        \State $S_c(M) \gets \sum_{c_d} \Gamma[M][c \!\to\! c_d]$ for each $M$ \Comment{change volume}
        \State Retain masks where $S_c(M) > \theta_m(c)$; require $\geq n_{\min}$ retained
        \State $\mathcal{R}.\text{add}\big(c,\;\text{argmax}_{M} S_c(M)\big)$ \Comment{peak interval}
    \EndFor
\EndIf
\end{algorithmic}
\end{algorithm}

\begin{algorithm}[!h]
\caption{Object-Level Task Construction}\label{alg:object_level}
\begin{algorithmic}[1]
\Require Preprocessed tile, task type $\tau \!\in\! \{\text{OP, BOC, OAM, OCM, OHM, OCO}\}$
\Ensure Object-level QA metadata $\mathcal{R}$
\Statex \textbf{Phase 1: Target Mask Selection}
\If{$\tau \in \{\text{OP, BOC, OCM}\}$}
    \State $\hat{\mathcal{M}} \gets \mathcal{M}_{\mathrm{inter}}$ \Comment{interval-specific transitions}
\Else
    \State $\hat{\mathcal{M}} \gets \{M_{\mathrm{total}}\}$ \Comment{cumulative end-state}
\EndIf
\Statex \textbf{Phase 2: Region Extraction}
\For{each mask $M \in \hat{\mathcal{M}}$, each target class $c$}
    \State Define binary mask $B$ per $\tau$:
    \State \quad OP:\; $B \!\gets\! \{p \mid c_d(v_p) \!=\! c\}$ \Comment{by destination}
    \State \quad BOC, OCM:\; $B \!\gets\! \{p \mid c_s(v_p) \!=\! c\}$ \Comment{by source}
    \State \quad OAM, OHM:\; $B \!\gets\! \{p \mid v_p \!=\! 10\,c_s\!+\!c_d\}$ \Comment{exact transition}
    \State \quad OCO:\; $B \!\gets\! \{p \mid c_d(v_p) \!\in\! \{4,5\}\}$ \Comment{buildings/playground}
    \State $\{R_i\} \gets \mathrm{CC}(B)$;\; discard $R_i$ if $\mathrm{Area}(R_i) < \theta_a(c)$ or $\mathrm{FR}(R_i) < \theta_f$
    \If{$\tau = \text{BOC}$}
        \State Collect all distinct $c_d$ values within each retained $R_i$
    \EndIf
\EndFor
\Statex \textbf{Phase 3: Temporal Localization} \Comment{memory \& ordering tasks}
\For{each retained region $R$}
    \If{$\tau \in \{\text{OAM},\, \text{OCO}\}$}
        \State Scan $\mathcal{M}_{\mathrm{inter}}$ backward; find earliest interval with dst-coverage $\geq \theta_{\mathrm{cov}}$
        \State Record $[t_0^{(j)}, t_1^{(j)}]$ as temporal anchor
    \ElsIf{$\tau = \text{OCM}$}
        \State Verify $\geq 2$ preceding intervals with no change in $R$
        \State Record current interval as change timestamp
    \ElsIf{$\tau = \text{OHM}$}
        \State Scan $\mathcal{M}_{\mathrm{inter}}$ chronologically over $R$'s pixels
        \State At each interval, identify dominant transition code
        \State Discard if non-dominant area $> 0.1 \times$ dominant area or $|\text{chain}| < 2$
    \EndIf
\EndFor
\Statex \textbf{Phase 4: Grounding}
\For{each retained region $R$}
    \State \textbf{Box}: compute horizontal bounding box of $R$
    \State \textbf{Point}: find interior pixel via distance transform; project to geo-coordinate
\EndFor
\end{algorithmic}
\end{algorithm}

Two tasks require additional post-processing beyond Algorithm~\ref{alg:object_level}. \textbf{Region Development Comparison~(RDC)} performs two separate region extractions on $M_{\mathrm{total}}$: one isolating building/playground destination clusters and another isolating vegetation/non-vegetated surface/water clusters. It then composes four-region combinations from the candidate pool, subject to three filtering criteria: \textbf{i)}~spatial non-containment, ensuring no selected region is enclosed by another; \textbf{ii)}~balanced area ratios, requiring $\max(\mathrm{Area}) / \min(\mathrm{Area}) \leq 4$ so that no region dominates the comparison by sheer size, and a coefficient of variation~(CV~$= \sigma / \mu$, the standard deviation of the four region areas divided by their mean) $\leq 0.7$ to further ensure that the four regions are of comparable scale; and \textbf{iii)}~fill-ratio discrimination constraints to guarantee a clear answer. The building-side cluster with the highest fill ratio is assigned as the most developed region. \textbf{Cross-Sequence Object Construction Ordering~(CSOCO)} reuses the outputs of OCO: it pools all retained building and playground objects across tiles into a global set, then pairs objects from different tiles whose construction intervals do not overlap. For each pair, the object with the earlier construction time is designated as the ground truth.

\subsection{Prompt Templates for All Sub-Tasks}\label{sec:templates}

After the rule-based programs extract the structured metadata records $\mathcal{R}$ described above, each record is converted into a QA pair by sampling one template from a predefined set of stylistically diverse prompts and populating all placeholders with concrete values from the metadata. To ensure linguistic variety and mitigate memorization of surface patterns, we design 5 templates for each grounding variant of each sub-task. The order of answer options is fully randomized across all questions to prevent positional bias. For tasks with both box-grounded and coordinate-grounded variants, the coordinate-grounded templates prepend a spatial reference preamble that specifies the coordinate reference system and image corner coordinates before the task-specific question.

The following placeholder notation is used across all templates:
\begin{itemize}
    \item \texttt{\{time\_0\}}, \texttt{\{time\_1\}}: observation timestamps.
    \item \texttt{\{land\_cover\}}, \texttt{\{land\_cover\_1\}}, \texttt{\{land\_cover\_2\}}: land-cover class names.
    \item \texttt{<\{HBB\_Box\}>}, \texttt{<\{HBB\_Box\_$N$\}>}: horizontal bounding boxes for spatial grounding.
    \item \texttt{\{crs\_system\}}, \texttt{\{crs\_system\_$N$\}}: coordinate reference system identifiers.
    \item \texttt{\{coo\_$N$\}}, \texttt{\{coo$N$\_$M$\}}: geographic coordinates for point-based spatial grounding.
\end{itemize}

\noindent The complete set of prompt templates for all 12 sub-tasks is presented below, organized by the four cognitive levels.

\paragraph{a. Land Cover Perception:}

\begin{templatebox}[Object Perception (OP)]{TemplLCP}
\textit{Box-Grounded:}
\begin{enumerate}\small
    \item Classify the land cover type of the <\{HBB\_Box\}>-grounded region in \{time\_0\}.
    \item In \{time\_0\}, what is the land cover type of the region grounded by <\{HBB\_Box\}>?
    \item Identify the land-cover type of the region grounded by <\{HBB\_Box\}> in \{time\_0\}.
    \item What land-cover class does the <\{HBB\_Box\}>-grounded region belong to at \{time\_0\}?
    \item Provide the land-cover classification for the area defined by <\{HBB\_Box\}> at \{time\_0\}.
\end{enumerate}

\textit{Coordinate-Grounded:}
\begin{enumerate}\small
    \item The coordinate reference system is \{crs\_system\}, the coordinate of the upper left corner of the image is \{coo\_1\}, the lower right is \{coo\_2\}. Classify the land cover type of the \{coo\_3\}-grounded region in \{time\_0\}.
    \item The coordinate reference system is \{crs\_system\}. The upper-left corner of the image is located at \{coo\_1\}, and the lower-right corner at \{coo\_2\}. In \{time\_0\}, what is the land cover type of the region grounded by coordinate \{coo\_3\}?
    \item The image is defined under the \{crs\_system\} coordinate reference system, with \{coo\_1\} as the upper-left coordinate and \{coo\_2\} as the lower-right. Identify the land-cover type of the region grounded by the coordinate \{coo\_3\} in \{time\_0\}.
    \item The coordinate reference system is \{crs\_system\}, the coordinate of the upper left corner of the image is \{coo\_1\}, the lower right is \{coo\_2\}. Now Answer: What land-cover class does the \{coo\_3\}-grounded region belong to at \{time\_0\}?
    \item Given that the image uses the \{crs\_system\} coordinate reference system, where \{coo\_1\} marks the upper-left corner and \{coo\_2\} the lower-right. Provide the land-cover classification for the area defined by the coordinate \{coo\_3\} at \{time\_0\}.
\end{enumerate}
\end{templatebox}

\paragraph{b. Temporal Recognition:}

\begin{templatebox}[Bi-Temporal Class-Level Change Recognition (BCC)]{TemplTR}
\begin{enumerate}\small
    \item What changes has the \{land\_cover\} undergone between \{time\_0\} and \{time\_1\}?
    \item How has the \{land\_cover\} changed comparing \{time\_0\} and \{time\_1\}?
    \item What changes can be observed in the \{land\_cover\} between \{time\_0\} and \{time\_1\}?
    \item Comparing \{time\_0\} vs.\ \{time\_1\}, what changes can be identified in the \{land\_cover\}?
    \item What is the two-point delta in the \{land\_cover\} from \{time\_0\} to \{time\_1\} (consider only these two timestamps)?
\end{enumerate}
\end{templatebox}

\begin{templatebox}[Bi-Temporal Class-Level Area Change Recognition (BCAC)]{TemplTR}
\begin{enumerate}\small
    \item How has the area of \{land\_cover\} changed between \{time\_0\} and \{time\_1\}? Choose from expanded or reduced.
    \item Classify the change in \{land\_cover\} area between \{time\_0\} and \{time\_1\}, pick expanded or reduced only.
    \item Determine how the area of \{land\_cover\} changed comparing \{time\_0\} and \{time\_1\}, your answer must be either expanded or reduced.
    \item How did the area of \{land\_cover\} change comparing \{time\_0\} vs.\ \{time\_1\}? Select one: expanded or reduced.
    \item Between \{time\_0\} and \{time\_1\}, did the area of \{land\_cover\} expand or reduce? Just pick one.
\end{enumerate}
\end{templatebox}

\begin{templatebox}[Long-Temporal Class-Level Change Recognition (LCC)]{TemplTR}
\begin{enumerate}\small
    \item What changes has the \{land\_cover\} undergone in the period of \{time\_0\} to \{time\_1\}?
    \item Across the entire time span from \{time\_0\} to \{time\_1\}, what changes did the \{land\_cover\} undergo?
    \item During the interval spanning \{time\_0\} to \{time\_1\}, what changes did the \{land\_cover\} experience?
    \item Over the timeframe from \{time\_0\} through \{time\_1\}, what changes are observed in the \{land\_cover\}?
    \item Within the period \{time\_0\}--\{time\_1\}, what changes has the \{land\_cover\} exhibited?
\end{enumerate}
\end{templatebox}

\begin{templatebox}[Bi-Temporal Object-Level Change Recognition (BOC)]{TemplTR}
\textit{Box-Grounded:}
\begin{enumerate}\small
    \item What changes have occurred to the land cover within the region <\{HBB\_BOX\}> (originally \{land\_cover\}) comparing \{time\_0\} and \{time\_1\}? Note that different parts of this region may have changed to different land cover types. Select ALL applicable transitions.
    \item Comparing \{time\_0\} and \{time\_1\}, what land cover transitions occurred within the region <\{HBB\_BOX\}> (originally \{land\_cover\})? Different areas within this region may have undergone different changes. Select ALL that apply.
    \item Examine the region <\{HBB\_BOX\}> (originally \{land\_cover\}) between \{time\_0\} and \{time\_1\}. What types of land cover changes occurred within this region? Multiple transitions may have happened in different parts of the region. Select ALL applicable changes.
    \item For the region <\{HBB\_BOX\}> (originally \{land\_cover\}), compare \{time\_0\} and \{time\_1\}: what land cover transitions took place? The region may contain multiple types of changes in different areas. Select ALL that apply.
    \item Between \{time\_0\} and \{time\_1\}, what land cover changes occurred within the region <\{HBB\_BOX\}> (originally \{land\_cover\})? Since this region may cover areas with different changes, select ALL applicable transitions.
\end{enumerate}

\textit{Coordinate-Grounded:}
\begin{enumerate}\small
    \item The coordinate reference system is \{crs\_system\}, the coordinate of the upper left corner of the image is \{coo\_1\}, the lower right is \{coo\_2\}. Now answer: What land cover changes have occurred at the location \{coo\_3\} (originally \{land\_cover\}) comparing \{time\_0\} and \{time\_1\}? The surrounding area may have undergone different types of transitions. Select ALL that apply.
    \item The coordinate reference system is \{crs\_system\}. The upper-left corner of the image is located at \{coo\_1\}, and the lower-right corner is at \{coo\_2\}. Now answer: Comparing \{time\_0\} and \{time\_1\}, what land cover transitions occurred at coordinate \{coo\_3\} (originally \{land\_cover\})? Multiple types of changes may have happened in this area. Select ALL applicable transitions.
    \item Using the coordinate reference system \{crs\_system\}, the image spans from \{coo\_1\} at the upper left to \{coo\_2\} at the lower right. Please answer: What types of land cover changes occurred at \{coo\_3\} (originally \{land\_cover\}) between \{time\_0\} and \{time\_1\}? Different parts of this area may have changed to different land cover types. Select ALL that apply.
    \item The image is defined under the coordinate reference system \{crs\_system\}, with \{coo\_1\} marking the upper-left corner and \{coo\_2\} marking the lower-right corner. Now respond: For the location \{coo\_3\} (originally \{land\_cover\}), what land cover transitions took place comparing \{time\_0\} and \{time\_1\}? The area may contain multiple types of changes. Select ALL that apply.
    \item Under the coordinate reference system \{crs\_system\}, the image's upper-left coordinate is \{coo\_1\} and its lower-right coordinate is \{coo\_2\}. Now answer the following: Between \{time\_0\} and \{time\_1\}, what land cover changes occurred at \{coo\_3\} (originally \{land\_cover\})? Since this area may cover regions with different changes, select ALL applicable transitions.
\end{enumerate}
\end{templatebox}

\paragraph{c. Long-Term Memory:}

\begin{templatebox}[Object Appear Memory (OAM)]{TemplLTM}
\textit{Box-Grounded:}
\begin{enumerate}\small
    \item During which timeframe did the current continuous presence of \{land\_cover\} in \{time\_0\} grounded by <\{HBB\_Box\}> begin?
    \item In which time window did the current continuous presence of the \{land\_cover\} in \{time\_0\} grounded by <\{HBB\_Box\}> begin?
    \item Within the temporal observations, at what interval did the <\{HBB\_Box\}>-grounded \{land\_cover\} in \{time\_0\} begin its current continuous presence?
    \item Identify the interval during which the <\{HBB\_Box\}>-grounded \{land\_cover\} in \{time\_0\} began its current continuous presence.
    \item At what time period did the <\{HBB\_Box\}>-grounded \{land\_cover\} observed in \{time\_0\} start to maintain its current continuous presence?
\end{enumerate}

\textit{Coordinate-Grounded:}
\begin{enumerate}\small
    \item The coordinate reference system is \{crs\_system\}, the coordinate of the upper left corner of the image is \{coo\_1\}, the lower right is \{coo\_2\}. During which timeframe did the current continuous presence of \{land\_cover\} in \{time\_0\} grounded by the coordinate \{coo\_3\} begin?
    \item The coordinate reference system is \{crs\_system\}. The upper-left corner of the image is located at \{coo\_1\}, and the lower-right corner is at \{coo\_2\}. Now answer: In which time window did the current continuous presence of the \{land\_cover\} in \{time\_0\} at coordinate \{coo\_3\} begin?
    \item Using the coordinate reference system \{crs\_system\}, the image spans from \{coo\_1\} at the upper left to \{coo\_2\} at the lower right. Within the temporal observations, at what interval did the coordinate \{coo\_3\}-grounded \{land\_cover\} in \{time\_0\} begin its current continuous presence?
    \item The image is defined under the coordinate reference system \{crs\_system\}, with \{coo\_1\} marking the upper-left corner and \{coo\_2\} marking the lower-right corner. Now respond: Identify the interval during which the coordinate \{coo\_3\}-grounded \{land\_cover\} in \{time\_0\} began its current continuous presence.
    \item Given that the image uses the \{crs\_system\} coordinate reference system, where \{coo\_1\} marks the upper-left corner and \{coo\_2\} the lower-right. At what time period did the \{coo\_3\}-grounded \{land\_cover\} observed in \{time\_0\} start to maintain its current continuous presence?
\end{enumerate}
\end{templatebox}

\begin{templatebox}[Object Change Memory (OCM)]{TemplLTM}
\textit{Box-Grounded:}
\begin{enumerate}\small
    \item At which exact timestamp does the object \{land\_cover\} in \{time\_0\} referred to by <\{HBB\_Box\}> undergo a change to other land cover type?
    \item Identify the interval during which the \{land\_cover\} in \{time\_0\} grounded by <\{HBB\_Box\}> changed to other land cover.
    \item What time range captures the change in land cover type of the object \{land\_cover\} in \{time\_0\} grounded by <\{HBB\_Box\}>?
    \item During what time interval does the object marked as \{land\_cover\} in \{time\_0\} and grounded by <\{HBB\_Box\}> undergo a transition to another land-cover class?
    \item Specify the timeframe in which the object \{land\_cover\} from \{time\_0\}, localized by <\{HBB\_Box\}>, shifts to a different land-cover type.
\end{enumerate}

\textit{Coordinate-Grounded:}
\begin{enumerate}\small
    \item The coordinate reference system is \{crs\_system\}, the coordinate of the upper left corner of the image is \{coo\_1\}, the lower right is \{coo\_2\}. At which exact timestamp does the object \{land\_cover\} in \{time\_0\} referred to by the coordinate \{coo\_3\} undergo a change to other land cover type?
    \item Using the coordinate reference system \{crs\_system\}, the image spans from \{coo\_1\} at the upper left to \{coo\_2\} at the lower right. Identify the interval during which the \{land\_cover\} in \{time\_0\} grounded by the coordinate \{coo\_3\} changed to other land cover.
    \item The coordinate reference system is \{crs\_system\}. The upper-left corner of the image is located at \{coo\_1\}, and the lower-right corner is at \{coo\_2\}. Now answer: What time range captures the change in land cover type of the object \{land\_cover\} in \{time\_0\} grounded by the coordinate \{coo\_3\}?
    \item The image is defined under the coordinate reference system \{crs\_system\}, with \{coo\_1\} marking the upper-left corner and \{coo\_2\} marking the lower-right corner. Now respond: During what time interval does the object marked as \{land\_cover\} in \{time\_0\} and grounded by the coordinate \{coo\_3\} undergo a transition to another land-cover class?
    \item Given that the image uses the \{crs\_system\} coordinate reference system, where \{coo\_1\} marks the upper-left corner and \{coo\_2\} the lower-right. Now specify the timeframe in which the object \{land\_cover\} from \{time\_0\}, localized by the coordinate \{coo\_3\}, shifts to a different land-cover type.
\end{enumerate}
\end{templatebox}

\begin{templatebox}[Object History Memory (OHM)]{TemplLTM}
\textit{Box-Grounded:}
\begin{enumerate}\small
    \item In chronological order, what distinct phases of land cover did the <\{HBB\_Box\}>-grounded \{land\_cover\} in \{time\_0\} go through?
    \item In chronological order, list the distinct land cover phases that the \{land\_cover\} area in \{time\_0\} indicated by <\{HBB\_Box\}> has undergone.
    \item Provide the chronological sequence of distinct land cover phases for the \{land\_cover\} in \{time\_0\} referred to by <\{HBB\_Box\}>.
    \item What is the chronological sequence of distinct land cover phases for the object \{land\_cover\} in \{time\_0\} identified by <\{HBB\_Box\}>?
    \item List, in chronological order, the distinct land-cover stages experienced by the <\{HBB\_Box\}>-grounded \{land\_cover\} observed in \{time\_0\}.
\end{enumerate}

\textit{Coordinate-Grounded:}
\begin{enumerate}\small
    \item The coordinate reference system is \{crs\_system\}, the coordinate of the upper left corner of the image is \{coo\_1\}, the lower right is \{coo\_2\}. Now answer: In chronological order, what distinct phases of land cover did the \{land\_cover\} grounded by the coordinate \{coo\_3\} in \{time\_0\} go through?
    \item The coordinate reference system is \{crs\_system\}. The upper-left corner of the image is located at \{coo\_1\}, and the lower-right corner at \{coo\_2\}. In chronological order, what distinct land-cover phases did the \{land\_cover\} grounded at coordinate \{coo\_3\} in \{time\_0\} undergo?
    \item The image is defined under the \{crs\_system\} coordinate reference system, with \{coo\_1\} as the upper-left coordinate and \{coo\_2\} as the lower-right. Provide the chronological sequence of distinct land cover phases for the \{land\_cover\} in \{time\_0\} referred to by the coordinate \{coo\_3\}.
    \item Given that the image uses the \{crs\_system\} coordinate reference system, where \{coo\_1\} marks the upper-left corner and \{coo\_2\} the lower-right. What is the chronological sequence of distinct land cover phases for the object \{land\_cover\} in \{time\_0\} identified by the coordinate \{coo\_3\}?
    \item Given that the image uses the \{crs\_system\} coordinate reference system, where \{coo\_1\} marks the upper-left corner and \{coo\_2\} the lower-right. List the distinct land-cover stages experienced by the \{coo\_3\}-grounded \{land\_cover\} observed in \{time\_0\} in chronological order.
\end{enumerate}
\end{templatebox}

\paragraph{d. Spatio-Temporal Reasoning:}

\begin{templatebox}[Class-Level Change Magnitude Estimation (CCME)]{TemplSTR}
\begin{enumerate}\small
    \item During which timeframe did the most substantial transformations in \{land\_cover\} occur?
    \item In which interval did the \{land\_cover\} undergo the sharpest change?
    \item Identify the time interval during which the \{land\_cover\} exhibited the greatest degree of change.
    \item What is the timeframe in which the \{land\_cover\} showed the most marked shift?
    \item Specify the interval over which the \{land\_cover\} experienced the most substantial changes.
\end{enumerate}
\end{templatebox}

\begin{templatebox}[Region Development Comparison (RDC)]{TemplSTR}
\begin{enumerate}\small
    \item Comparing Region1 <\{HBB\_Box\_1\}>, Region2 <\{HBB\_Box\_2\}>, Region3 <\{HBB\_Box\_3\}> and Region4 <\{HBB\_Box\_4\}>, which is developed more from \{time\_0\} to \{time\_1\}?
    \item Among Region1 <\{HBB\_Box\_1\}>, Region2 <\{HBB\_Box\_2\}>, Region3 <\{HBB\_Box\_3\}> and Region4 <\{HBB\_Box\_4\}>, which is the most developed from \{time\_0\} to \{time\_1\}?
    \item Between Region1 <\{HBB\_Box\_1\}>, Region2 <\{HBB\_Box\_2\}>, Region3 <\{HBB\_Box\_3\}>, and Region4 <\{HBB\_Box\_4\}>, which one shows greater development from \{time\_0\} to \{time\_1\}?
    \item Considering Region1 <\{HBB\_Box\_1\}>, Region2 <\{HBB\_Box\_2\}>, Region3 <\{HBB\_Box\_3\}>, and Region4 <\{HBB\_Box\_4\}>, identify which region is the most developed between \{time\_0\} and \{time\_1\}.
    \item From \{time\_0\} to \{time\_1\}, which of the regions: Region1 <\{HBB\_Box\_1\}>, Region2 <\{HBB\_Box\_2\}>, Region3 <\{HBB\_Box\_3\}>, or Region4 <\{HBB\_Box\_4\}>, demonstrates the most substantial development?
\end{enumerate}
\end{templatebox}

\begin{templatebox}[Object Construction Ordering (OCO)]{TemplSTR}
\textit{Box-Grounded:}
\begin{enumerate}\small
    \item Comparing the \{land\_cover\_1\} grounded by <\{HBB\_Box\_1\}> and the \{land\_cover\_2\} grounded by <\{HBB\_Box\_2\}> in \{time\_0\}, which is built earlier?
    \item Comparing the \{land\_cover\_1\} grounded by <\{HBB\_Box\_1\}> with the \{land\_cover\_2\} grounded by <\{HBB\_Box\_2\}> in \{time\_0\}, identify which was built earlier.
    \item Which was built earlier: the \{land\_cover\_1\} at <\{HBB\_Box\_1\}> or the \{land\_cover\_2\} at <\{HBB\_Box\_2\}> in \{time\_0\}?
    \item Of the \{land\_cover\_1\} grounded by <\{HBB\_Box\_1\}> and the \{land\_cover\_2\} grounded by <\{HBB\_Box\_2\}> in \{time\_0\}, determine which appeared first in time.
    \item Between the \{land\_cover\_1\} located at <\{HBB\_Box\_1\}> and the \{land\_cover\_2\} located at <\{HBB\_Box\_2\}> in \{time\_0\}, which one was constructed first?
\end{enumerate}

\textit{Coordinate-Grounded:}
\begin{enumerate}\small
    \item The coordinate reference system is \{crs\_system\}, the coordinate of the upper left corner of the image is \{coo\_1\}, the lower right is \{coo\_2\}. Now answer: Comparing the \{land\_cover\_1\} grounded by the coordinate \{coo\_3\} and the \{land\_cover\_2\} grounded by \{coo\_4\}, which is built earlier?
    \item The coordinate reference system is \{crs\_system\}, with \{coo\_1\} as the upper-left corner and \{coo\_2\} as the lower-right corner. Now answer: Between the \{land\_cover\_1\} located at \{coo\_3\} and the \{land\_cover\_2\} located at \{coo\_4\} in \{time\_0\}, which one was built earlier?
    \item Given the coordinate reference system \{crs\_system\}, with \{coo\_1\} defining the upper-left boundary and \{coo\_2\} the lower-right, identify which feature was built earlier: the \{land\_cover\_1\} grounded at \{coo\_3\} or the \{land\_cover\_2\} grounded at \{coo\_4\} in \{time\_0\}.
    \item The coordinate reference system is \{crs\_system\}, the coordinate of the upper left corner of the image is \{coo\_1\}, the lower right is \{coo\_2\}. Comparing the \{land\_cover\_1\} grounded by \{coo\_3\} with the \{land\_cover\_2\} grounded by \{coo\_4\} in \{time\_0\}, identify which was built earlier.
    \item Given that the image uses the \{crs\_system\} coordinate reference system, where \{coo\_1\} marks the upper-left corner and \{coo\_2\} the lower-right. Between the \{land\_cover\_1\} located at \{coo\_3\} and the \{land\_cover\_2\} located at \{coo\_4\} in \{time\_0\}, which one was constructed first?
\end{enumerate}
\end{templatebox}

\begin{templatebox}[Cross-Sequence Object Construction Ordering (CSOCO)]{TemplSTR}
\textit{Box-Grounded:}
\begin{enumerate}\small
    \item Comparing the \{land\_cover\_1\} grounded by <\{HBB\_Box\_1\}> in Region\#1 in \{time\_0\} and the \{land\_cover\_2\} grounded by <\{HBB\_Box\_2\}> in Region\#2 in \{time\_1\}, which is built earlier?
    \item Between the \{land\_cover\_1\} grounded by <\{HBB\_Box\_1\}> in Region\#1 in \{time\_0\} and the \{land\_cover\_2\} grounded by <\{HBB\_Box\_2\}> in Region\#2 in \{time\_1\}, which one was built earlier?
    \item Comparing the \{land\_cover\_1\} located at <\{HBB\_Box\_1\}> in Region\#1 in \{time\_0\} with the \{land\_cover\_2\} located at <\{HBB\_Box\_2\}> in Region\#2 in \{time\_1\}, identify which was constructed first.
    \item Comparing the \{land\_cover\_1\} located at <\{HBB\_Box\_1\}> in Region\#1 in \{time\_0\} with the \{land\_cover\_2\} located at <\{HBB\_Box\_2\}> in Region\#2 in \{time\_1\}, determine which appeared first in time.
    \item Of the \{land\_cover\_1\} grounded by <\{HBB\_Box\_1\}> in Region\#1 in \{time\_0\} and the \{land\_cover\_2\} grounded by <\{HBB\_Box\_2\}> in Region\#2 in \{time\_1\}, which one was constructed first?
\end{enumerate}

\textit{Coordinate-Grounded:}
\begin{enumerate}\small
    \item The coordinate reference system, upper left and lower right corner coordinate of Region\#1 and Region\#2 are respectively: Region\#1: \{crs\_system\_1\}, \{coo1\_1\}, \{coo2\_1\}; Region\#2: \{crs\_system\_2\}, \{coo1\_2\}, \{coo2\_2\}. Now answer: Comparing the \{land\_cover\_1\} grounded by coordinate \{coo\_3\} in Region\#1 in \{time\_0\} and the \{land\_cover\_2\} grounded by \{coo\_4\} in Region\#2 in \{time\_1\}, which is built earlier?
    \item Region\#1 uses \{crs\_system\_1\} with coordinates spanning from \{coo1\_1\} (upper left) to \{coo2\_1\} (lower right). Region\#2 uses \{crs\_system\_2\} with coordinates from \{coo1\_2\} to \{coo2\_2\}. Now answer: Between the \{land\_cover\_1\} grounded by \{coo\_3\} in Region\#1 in \{time\_0\} and the \{land\_cover\_2\} grounded by \{coo\_4\} in Region\#2 in \{time\_1\}, which one was built earlier?
    \item The coordinate reference system, upper left and lower right corner coordinate of Region\#1 and Region\#2 are respectively: Region\#1: \{crs\_system\_1\}, \{coo1\_1\}, \{coo2\_1\}; Region\#2: \{crs\_system\_2\}, \{coo1\_2\}, \{coo2\_2\}. Answer: Comparing the \{land\_cover\_1\} located at \{coo\_3\} in Region\#1 in \{time\_0\} with the \{land\_cover\_2\} located at \{coo\_4\} in Region\#2 in \{time\_1\}, identify which was constructed first.
    \item Region\#1 uses \{crs\_system\_1\} with coordinates spanning from \{coo1\_1\} (upper left) to \{coo2\_1\} (lower right). Region\#2 uses \{crs\_system\_2\} with coordinates from \{coo1\_2\} to \{coo2\_2\}. Comparing the \{land\_cover\_1\} located at \{coo\_3\} in Region\#1 in \{time\_0\} with the \{land\_cover\_2\} located at \{coo\_4\} in Region\#2 in \{time\_1\}, determine which appeared first in time.
    \item The coordinate reference system, upper left and lower right corner coordinate of Region\#1 and Region\#2 are respectively: Region\#1: \{crs\_system\_1\}, \{coo1\_1\}, \{coo2\_1\}; Region\#2: \{crs\_system\_2\}, \{coo1\_2\}, \{coo2\_2\}. Of the \{land\_cover\_1\} grounded by the coordinate \{coo\_3\} in Region\#1 in \{time\_0\} and the \{land\_cover\_2\} grounded by the coordinate \{coo\_4\} in Region\#2 in \{time\_1\}, which one was constructed first?
\end{enumerate}
\end{templatebox}

\subsection{Natural-Language Response Generation for ChronoInstruct}\label{sec:nl_response}

The pipeline described above produces structured QA pairs in multiple-choice format, which serve both the ChronoBench evaluation and the ChronoInstruct instruction-tuning dataset. To enrich the model's output repertoire during instruction tuning, ChronoInstruct provides each QA pair in three complementary answer formats: \textbf{i)}~\textit{multiple-choice}, which retains the original option-letter answer; \textbf{ii)}~\textit{short-text}, which replaces the option letter(s) with the corresponding semantic content~(e.g., ``B.''~$\to$~``vegetation''), providing a concise factual answer without candidate options; and \textbf{iii)}~\textit{natural-language}, a fluent, self-contained English response that rephrases the answer into complete sentences. Among these, the multiple-choice and short-text formats are directly derived from the rule-based extraction outputs without additional processing, whereas the natural-language format requires an additional generation stage. This subsection describes the construction of natural-language responses.

We employ Gemini-3-Flash~\cite{team2023gemini} to transform each multiple-choice QA pair into a fluent, self-contained English response. At inference time, the model receives only the textual components of the QA pair without any visual input. A carefully designed system prompt governs the generation process by enforcing six key constraints: \textbf{i)}~strict factual fidelity to the ground-truth answer without inventing additional details; \textbf{ii)}~omission of all spatial references such as bounding-box coordinates, geographic coordinates, and coordinate reference systems; \textbf{iii)}~expressive variety in vocabulary and sentence structure to avoid formulaic patterns; \textbf{iv)}~task-specific handling strategies tailored to different answer types, including ordered sequences, unordered multi-selections, binary judgments, and comparisons; \textbf{v)}~brevity within 1--3 sentences; and \textbf{vi)}~clean output without any preamble or meta-commentary. The complete system prompt is presented below.

\begin{systempromptbox}[System Prompt for Natural-Language Response Generation]
\small
You are an expert in remote sensing temporal image analysis.

\medskip
\noindent\textbf{Land cover categories:}
\begin{enumerate}[nosep, leftmargin=1.5em]
    \item \textbf{vegetation} --- areas covered by plants, grass, trees, or crops
    \item \textbf{buildings} --- man-made structures such as houses, factories, or commercial buildings
    \item \textbf{non-vegetated surface} --- bare soil, roads, parking lots, or other exposed ground
    \item \textbf{water} --- rivers, lakes, ponds, or other water bodies
    \item \textbf{playground} --- sports fields, recreational areas, or athletic facilities
\end{enumerate}

\medskip
\noindent Your task is to convert a multiple-choice letter answer into a fluent, natural English response, as if you were directly answering a remote sensing analysis question without any options presented.

\medskip
\noindent\textbf{Rules:}
\begin{enumerate}[nosep, leftmargin=1.5em]
    \item \textsc{Factual Fidelity}: Convey exactly the same factual content as the letter answer. You MUST NOT invent, infer, or add any detail that is not present in the given answer and options. Do not speculate about causes, consequences, or spatial extent.

    \item \textsc{Omit Spatial References}: The question may mention bounding box coordinates (e.g., $\langle$112, 409, 205, 437$\rangle$), geographic coordinates (e.g., (661778.5, 5188361.5)), coordinate reference systems (e.g., EPSG:26914), or corner coordinates. Do NOT repeat any of these in your answer. Refer to objects using only their semantic labels (e.g., ``the building'', ``this location'', ``Region1'') without restating numeric coordinates or bounding boxes.

    \item \textsc{Expressive Variety}: Use diverse sentence structures, vocabulary, and phrasing across samples. Vary openings, connectors, and word choices. Avoid formulaic or repetitive patterns. For example:
    \begin{itemize}[nosep, leftmargin=1em]
        \item Instead of always ``The land cover is X'', also use ``This area is covered by X'', ``X occupies this location'', ``The surface here consists of X'', etc.
        \item Instead of always ``changed from X to Y'', also use ``shifted from X to Y'', ``was converted from X into Y'', ``underwent a transition from X to Y'', etc.
    \end{itemize}

    \item \textsc{Task-Specific Handling}:
    \begin{itemize}[nosep, leftmargin=1em]
        \item \textit{Ordered Sequence (History Memory)}: The letter order represents a chronological sequence of land cover phases. Your answer MUST preserve this exact temporal ordering. Vary the phrasing, e.g., ``transitioned from X to Y and subsequently to Z'', ``progressed through phases: first X, then Y, finally Z'', ``began as X, later became Y, and ultimately turned into Z''.
        \item \textit{Unordered Multiple Selections}: List all detected items naturally. Vary how you enumerate them.
        \item \textit{Single Selection}: Give a direct, concise response.
        \item \textit{Binary (Expanded/Reduced)}: Use varied vocabulary (e.g., ``increased/grew/broadened/expanded'' or ``diminished/contracted/shrunk/declined'').
        \item \textit{Comparison}: State the comparison result directly using the object's label only.
        \item \textit{Time Period}: Weave the time period naturally into the sentence.
    \end{itemize}

    \item \textsc{Brevity}: 1--3 sentences maximum. No filler, no hedging, no caveats.

    \item \textsc{Output}: Only the natural language answer itself. No preamble like ``The answer is'', no quotes, no explanation.
\end{enumerate}
\end{systempromptbox}

Table~\ref{tab:instruct_examples} presents representative examples from ChronoInstruct across all 12 sub-tasks, illustrating the three answer formats: the original multiple-choice letter answer, the short-text answer that replaces option letters with their semantic content, and the natural-language response generated by the pipeline described above. The three formats differ not only in their answers but also in their user-prompt construction. For the \textit{multiple-choice} format, the question is followed by a ``Choose from: A.~$\langle$option$\rangle$, B.~$\langle$option$\rangle$, $\ldots$'' enumeration of candidate options. For the \textit{short-text} format, the candidate options are removed and replaced by a randomly sampled brevity suffix such as ``\textit{Answer concisely.}'' or ``\textit{List all applicable answers briefly.}'', depending on whether the task expects a single answer or multiple answers. For the \textit{natural-language} format, the user prompt contains only the bare question without any options or suffix.

\begin{table*}[!htb]
\centering
\caption{Representative ChronoInstruct examples across all 12 sub-tasks, showing the question alongside three answer formats: multiple-choice option letter(s), short-text answer, and natural-language response. Spatial coordinates are abbreviated with ``$[\cdots]$'' for compactness.}
\label{tab:instruct_examples}
\setlength{\tabcolsep}{4pt}
\renewcommand{\arraystretch}{1.15}
\resizebox{\textwidth}{!}{
\begin{tabular}{p{1.2cm} p{6.5cm} c p{3.8cm} p{5.5cm}}
\toprule
\textbf{Task} & \textbf{Question} & \textbf{Option} & \textbf{Short-Text Answer} & \textbf{Natural-Language Response} \\
\midrule
\multicolumn{5}{l}{\cellcolor{LCPColor}\textit{Land Cover Perception}} \\
OP & In 2013, what is the land cover type of the region grounded by $\langle$580, 256, 769, 611$\rangle$? & D. & vegetation & In 2013, the region was covered by vegetation. \\
\midrule
\multicolumn{5}{l}{\cellcolor{TAColor}\textit{Temporal Recognition}} \\
BCC & What is the two-point delta in the non-vegetated surface from 2021 to 2023 (consider only these two timestamps)? & D. & non-vegetated surface to vegetation & From 2021 to 2023, the non-vegetated surface transitioned to vegetation. \\
\addlinespace
BCAC & How did the area of vegetation change comparing 2005 vs.\ 2023? Select one: expanded or reduced. & B. & Reduced & The area covered by vegetation diminished between 2005 and 2023. \\
\addlinespace
BOC & Comparing 2018 and 2022, what land cover transitions occurred within the region $\langle$526, 133, 554, 151$\rangle$ (originally building)? $[\cdots]$ Select ALL that apply. & A. & buildings to non-vegetated surface & From 2018 to 2022, the land cover in this area transitioned from buildings to non-vegetated surface. \\
\addlinespace
LCC & What changes has the non-vegetated surface undergone in the period of 2010 to 2022? & C. & non-vegetated surface to vegetation & From 2010 to 2022, the non-vegetated surface experienced transitions into vegetation. \\
\midrule
\multicolumn{5}{l}{\cellcolor{LTMColor}\textit{Long-Term Memory}} \\
OAM & Identify the interval during which the $\langle$0, 123, 181, 326$\rangle$-grounded vegetation in 2021 began its current continuous presence. & C. & 2011--2013 & The vegetation present at this location in 2021 has been continuously there since the period 2011--2013. \\
\addlinespace
OCM & Identify the interval during which the vegetation in 2008 grounded by $\langle$196, 791, 344, 908$\rangle$ changed to other land cover. & B. & 2015--2017 & The vegetation at this location transitioned to a different land cover type during the period 2015--2017. \\
\addlinespace
OHM & Provide the chronological sequence of distinct land cover phases for the building in 2023 referred to by $\langle$210, 155, 277, 172$\rangle$. & D, A. & vegetation $\to$ buildings & This location was originally covered by vegetation and later developed into a building. \\
\midrule
\multicolumn{5}{l}{\cellcolor{STRColor}\textit{Spatio-Temporal Reasoning}} \\
CCME & In which interval did the vegetation undergo the sharpest change? & B. & 2012--2014 & The vegetation experienced the sharpest change during the interval 2012--2014. \\
\addlinespace
RDC & Among Region1 $\langle$58, 19, 130, 293$\rangle$, Region2 $\langle$63, 465, 148, 647$\rangle$, Region3 $\langle$589, 130, 640, 321$\rangle$ and Region4 $\langle$700, 516, 862, 672$\rangle$, which is the most developed from 2010 to 2021? & C. & Region3 & Region3 showed the highest level of development between 2010 and 2021. \\
\addlinespace
OCO & Comparing the building\#1 grounded by $\langle$859, 200, 995, 278$\rangle$ and the building\#2 grounded by $\langle$0, 478, 96, 612$\rangle$ in 2023, which is built earlier? & B. & building\#2 & Building\#2 was constructed earlier than building\#1. \\
\addlinespace
CSOCO & Comparing the building\#1 at $\langle$838, 79, 913, 163$\rangle$ in Region\#1 in 2021 with the building\#2 at $\langle$33, 943, 131, 999$\rangle$ in Region\#2 in 2023, identify which was constructed first. & A. & building\#1 & Building\#1 in Region\#1 was constructed before building\#2 in Region\#2. \\
\bottomrule
\end{tabular}
}
\end{table*}
\vspace{-1em}

\section{Statistical Analysis of ChronoBench and ChronoInstruct}\label{sec:more_stat}

In this section, we present a more detailed statistical breakdown along three complementary dimensions: \textbf{i)}~sub-task QA distribution, \textbf{ii)}~temporal sequence length, and \textbf{iii)}~answer format composition. Together, these analyses offer a finer-grained view of the dataset characteristics that underpin the benchmark's evaluation coverage and the instruction-tuning dataset's training diversity.

\subsection{Per-Task QA Distribution}\label{sec:stat_task}

\begin{figure}[htb]
    \centering
    \includegraphics[width=\linewidth]{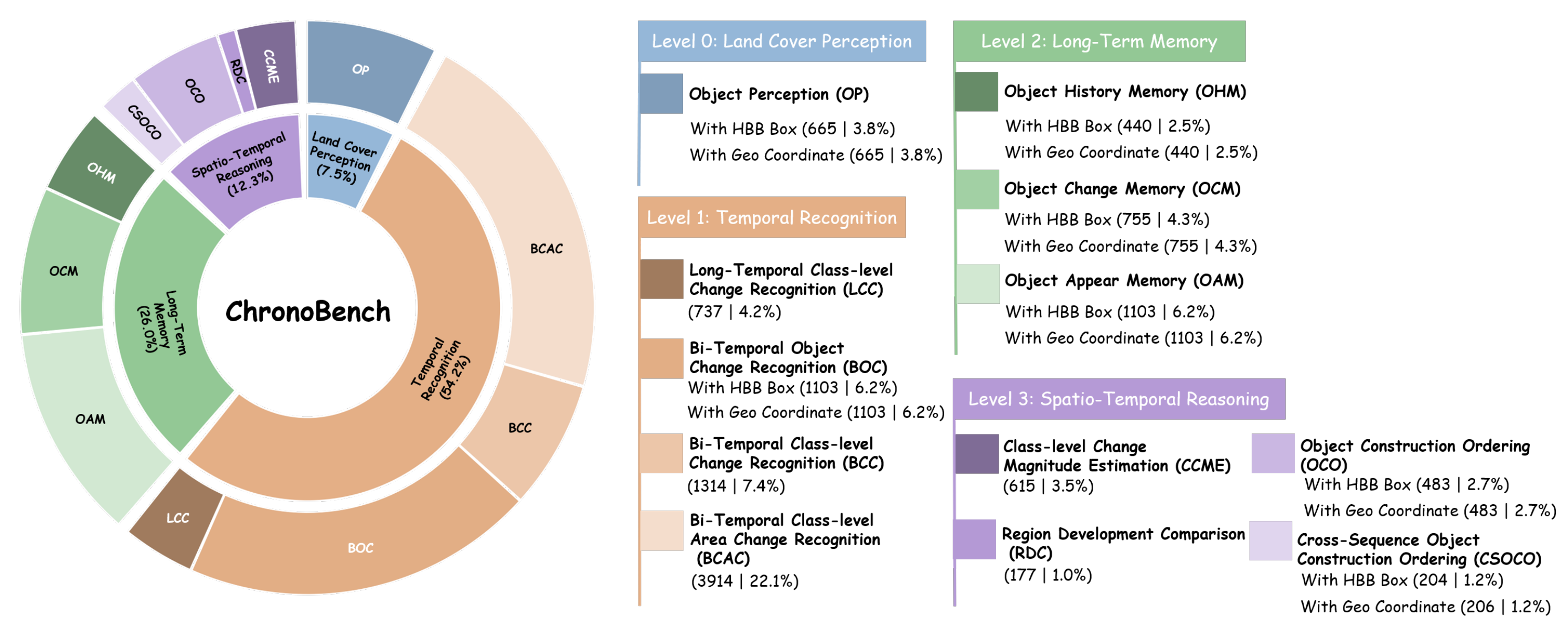}
    \caption{Hierarchical distribution of QA pairs across the four cognitive levels and 12 sub-tasks in \textbf{ChronoBench}~(17,689 pairs). The inner ring shows level-wise proportions; the outer ring shows the per-task breakdown, with spatially grounded tasks further split into HBB-box and geo-coordinate variants.}
    \label{fig:stat_bench}
\end{figure}

\begin{figure}[htb]
    \centering
    \includegraphics[width=\linewidth]{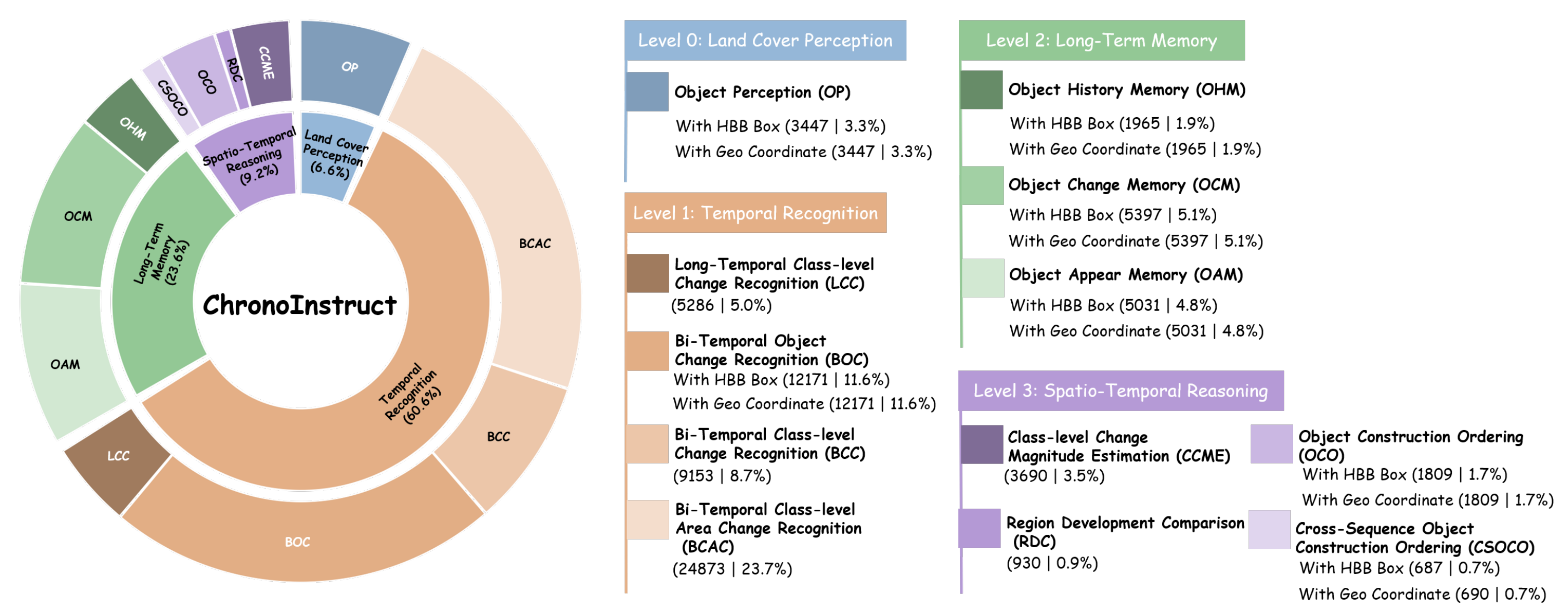}
    \caption{Hierarchical distribution of QA pairs across the four cognitive levels and 12 sub-tasks in \textbf{ChronoInstruct}~(104,949 pairs). Layout and color coding follow Figure~\ref{fig:stat_bench}.}
    \label{fig:stat_instruct}
\end{figure}

Figures~\ref{fig:stat_bench} and~\ref{fig:stat_instruct} present the hierarchical distribution of QA pairs across the four cognitive levels and their constituent sub-tasks for ChronoBench and ChronoInstruct, respectively.

In ChronoBench, Temporal Recognition dominates with 54.2\% of all QA pairs. Within this level, Bi-Temporal Class-Level Area Change Recognition~(BCAC) alone accounts for 22.1\%~(3,914 pairs), as every valid class--interval combination yields a binary expanded/reduced question. Bi-Temporal Object-Level Change Recognition~(BOC) contributes another 12.4\% across its two grounding variants~(1,103 pairs each). Long-Term Memory tasks constitute 26.0\%, with Object Appear Memory~(OAM) being the most prevalent~(12.4\%). Spatio-Temporal Reasoning, the most cognitively demanding level, accounts for 12.3\%, where the cross-sequence tasks~(CSOCO) are deliberately kept compact~(2.4\%) to reflect the inherently smaller pool of valid cross-tile comparison pairs. Land Cover Perception~(OP) occupies the remaining 7.5\%, evenly split between the two grounding modalities.

ChronoInstruct comprises 104,949 QA pairs derived from the DVL-Suite~\cite{xuan2025dynamicvl} training split~(1,534 image sequences). Temporal Recognition accounts for 60.5\% of the dataset, with BCAC~(24,873 pairs, 23.7\%) and BOC~(24,342 pairs across both grounding variants, 23.2\%) as the two largest sub-tasks. Long-Term Memory contributes 23.1\%, led by OCM~(10,794 pairs, 10.2\%) and OAM~(10,062 pairs, 9.6\%). Spatio-Temporal Reasoning constitutes 9.2\%, where CCME~(3,690 pairs, 3.5\%) is the dominant sub-task, while the cross-sequence tasks~(CSOCO) account for 1.4\% due to the inherently smaller pool of valid cross-tile comparison pairs. Land Cover Perception~(OP) makes up the remaining 6.6\%.

For most spatially grounded sub-tasks~(OP, BOC, OAM, OCM, OHM, OCO), the HBB-box and geo-coordinate variants are generated in equal numbers, ensuring balanced coverage of both grounding modalities. The exception is CSOCO, where the two variants may differ slightly~(e.g., 204 vs.\ 206 in ChronoBench, 687 vs.\ 690 in ChronoInstruct), because the cross-sequence pairing procedure randomly samples candidate sequences, leading to minor stochastic variation between the two grounding modes.

\subsection{Temporal Sequence Length}\label{sec:stat_length}

\begin{figure}[htb]
    \centering
    \includegraphics[width=1\linewidth]{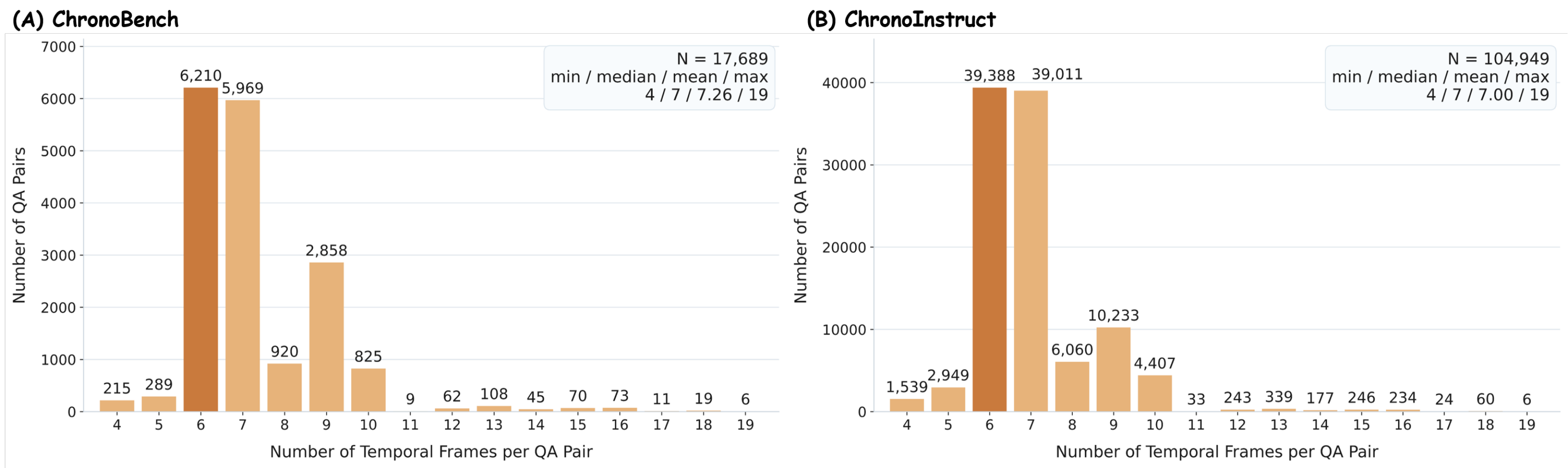}
    \caption{Distribution of the number of temporal frames per QA pair in \textbf{(A)}~ChronoBench and \textbf{(B)}~ChronoInstruct. Annotations above each bar indicate the QA pair count; summary statistics~(min, median, mean, max) are shown in the upper-right inset.}
    \label{fig:data_length}
\end{figure}

Figure~\ref{fig:data_length} shows the distribution of the number of temporal frames per QA pair for both datasets.

In ChronoBench~(Figure~\ref{fig:data_length}A), the frame count ranges from 4 to 19, with a median of 7 and a mean of 7.26. The distribution is strongly concentrated at 6 and 7 frames~(6,210 and 5,969 pairs, respectively), which together account for approximately 68.9\% of all QA pairs. This concentration reflects the typical number of available NAIP acquisition epochs across most U.S. regions in the DynamicVL dataset. Beyond 8 frames, the counts taper off into a long tail extending to 19 frames, capturing geographic tiles with denser temporal coverage. The maximum frame count of 19 arises from cross-sequence sub-tasks~(e.g., CSOCO), where two independent image sequences are concatenated into a single visual input, thereby extending the temporal span compared to single-sequence tasks.

ChronoInstruct~(Figure~\ref{fig:data_length}B) spans the same 4--19 range, with 104,949 pairs, a median of 7, and a mean of 7.00. The 6- and 7-frame bins jointly cover 74.7\% of the dataset~(39,388 and 39,011 pairs, respectively). The extended tail~(11--19 frames) is sparsely populated, confirming that ultra-long sequences remain rare but are nevertheless represented in both datasets to exercise the model's capacity for extended temporal reasoning.

\subsection{Answer Format Composition}\label{sec:stat_format}

\begin{figure}[htb]
    \centering
    \includegraphics[width=0.8\linewidth]{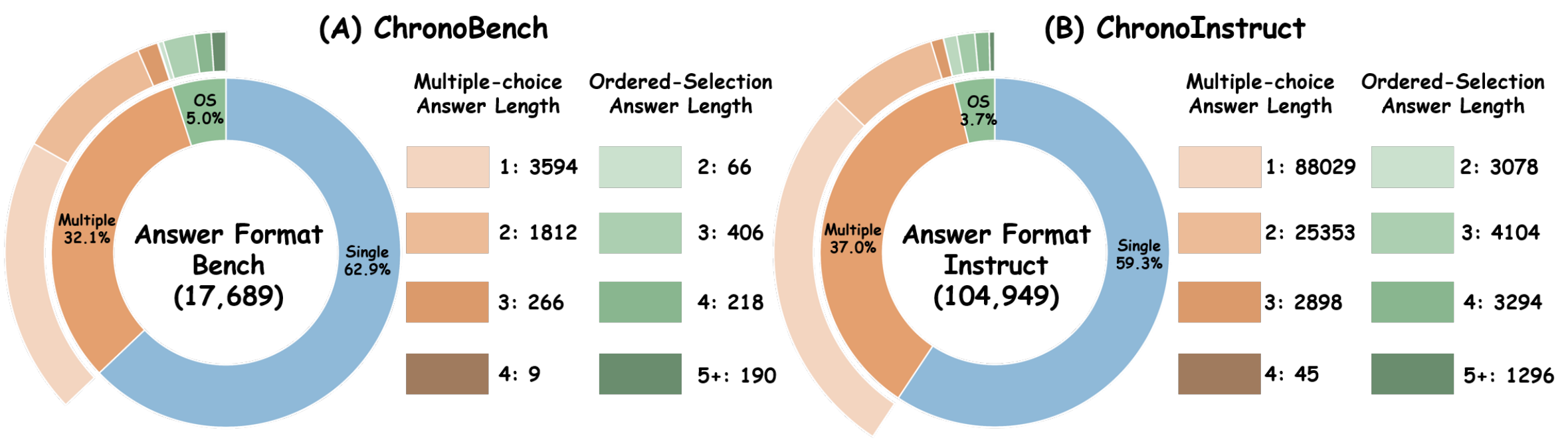}
    \caption{Answer format composition of \textbf{(A)}~ChronoBench and \textbf{(B)}~ChronoInstruct. The donut charts classify all QA pairs by the logical structure of their ground-truth answers: single~(exactly one correct option), multiple~(two or more correct options), and ordered-sequence~(OS, a chronologically ordered chain). For ChronoInstruct, each QA pair appears in three response formats~(multiple-choice, short-text, and natural-language), and the counts are aggregated across all three formats. Stacked bars further break down the number of correct options for the multiple category~(left) and the chain length for the OS category~(right).}
    \label{fig:qa_format}
\end{figure}

Figure~\ref{fig:qa_format} dissects the answer format characteristics of both datasets.

In ChronoBench~(Figure~\ref{fig:qa_format}A), single-choice questions represent the majority~(62.9\%), covering tasks with a unique correct answer such as OP, BCAC, CCME, RDC, and OCO/CSOCO. Multiple-choice questions account for 32.1\%, arising from tasks where multiple transitions may co-occur~(e.g., BCC, LCC, BOC). Among these, questions with a single correct option are the most common~(3,594), while questions requiring the model to identify two, three, or four simultaneous transitions number 1,812, 266, and 9, respectively. The ordered-selection~(OS) format is reserved exclusively for Object History Memory~(OHM), contributing 5.0\% of all questions. The majority of OS questions involve sequences of three land-cover phases~(406 pairs), with two-phase~(66), four-phase~(218), and five-or-more-phase sequences~(190) also present, reflecting the varying complexity of evolution histories across different geographic regions.

For ChronoInstruct~(Figure~\ref{fig:qa_format}B), it is important to note that every QA pair is provided in three complementary response formats~(i.e., multiple-choice, short-text, and free-form natural-language) so that the model is exposed to diverse answer styles during training. The statistics in Figure~\ref{fig:qa_format}B classify all 104,949 samples by the \emph{logical} answer structure (i.e., the number of ground-truth options involved), aggregating across the three response formats. Under this view, 59.3\% of the samples correspond to a single correct option, 37.0\% involve multiple correct options, and 3.7\% are ordered sequences. Within the multi-option category, single-correct-option instances dominate~(88,029), while those with two or more correct options total 28,296. For ordered-selection samples, three-phase and four-phase chains are the most frequent~(4,104 and 3,294, respectively), providing rich supervision for training the model to produce chronologically faithful outputs across all response formats.

\section{Experiment Setups}\label{sec:exp_setup}

This section provides the detailed experimental protocol, covering the benchmark evaluation pipeline~(Section~\ref{sec:eval_settings}) and the GeoChrono training configuration ~(Section~\ref{sec:train_settings}).

\subsection{Evaluation Settings}\label{sec:eval_settings}

As described in the main text, we evaluate three categories of models on ChronoBench: commercial MLLMs~(Gemini-3-Flash~\cite{team2023gemini}, Seed-1.6-Vision~\cite{bytedance2026seed16}, GPT-5.4~\cite{openai2025gpt41}), open-source general MLLMs~(InternVL-3.5 series~\cite{wang2025internvl3}, Qwen3-VL series~\cite{bai2025qwen3}), and remote sensing domain models~(TEOChat~\cite{irvin2024teochat}, EarthDial~\cite{soni2025earthdial}, DVLChat~\cite{xuan2025dynamicvl}). Since the official model weights of DVLChat are not publicly available, we reproduce this baseline by fine-tuning Qwen3-VL-4B-Instruct on the open-source DVL-Instruct training dataset released by the DynamicVL team~\cite{xuan2025dynamicvl}. Open-source models are deployed locally with Flash-Attention-2~\cite{dao2023flashattention2} and \texttt{bfloat16} precision; greedy decoding is adopted with \texttt{max\_new\_tokens\,=\,128}. Note that all Qwen3-VL variants used in this study refer to their Instruct
versions. Commercial models are accessed through their respective APIs, with the vision detail level set to the highest available setting and the reasoning effort set to \texttt{high}. All models share a unified prompt organization; the only difference in visual input is that locally deployed models receive images at their original resolution, whereas API-accessed models receive images preprocessed to meet the API's maximum-resolution constraint. In the following, we detail the prompt design and evaluation metrics.

\subsubsection{System Prompt Design}\label{sec:sys_prompt_design}

Each of the 12 benchmark sub-tasks is assigned a deterministic, task-specific system prompt. Despite per-task wording differences, all system prompts are assembled from five standardized components:

\begin{enumerate}[leftmargin=1.5em]
    \item \textbf{Expert role declaration.} The prompt assigns the model an expert role in remote sensing image analysis, with emphasis that varies by tasks. These tasks include land-cover classification for perception, temporal change detection for recognition, and temporal reasoning for memory and spatio-temporal tasks.

    \item \textbf{Input-format specification.} The prompt explicitly states the expected multimodal input layout. For single-sequence tasks, this takes the form:
    \begin{quote}\small\ttfamily
    YYYY: [image]\\
    YYYY: [image]\\
    \ldots
    \end{quote}
    For cross-sequence tasks~(CSOCO), the prompt specifies two region-specific sub-sequences:
    \begin{quote}\small\ttfamily
    Region\#1:\\
    YYYY: [image]\\
    \ldots\\[2pt]
    Region\#2:\\
    YYYY: [image]\\
    \ldots
    \end{quote}

    \item \textbf{Grounding specification.} Object-centric tasks include one of two grounding descriptions: \textit{(a)}~HBB-based grounding, where objects are localized by horizontal bounding boxes in normalized image coordinates $[0, 999]$; or \textit{(b)}~coordinate-based grounding, where objects are localized by geographic coordinates under a stated coordinate reference system. Class-level tasks omit this component.

    \item \textbf{Land-cover taxonomy.} All prompts explicitly enumerate the five land-cover categories~(vegetation, buildings, non-vegetated surface, water, playground) to ensure a consistent vocabulary between the system instruction and the model's output.

    \item \textbf{Task instruction and response constraint.} The final component describes the reasoning objective and constrains the output format to one of three serialization modes:
    \begin{itemize}[nosep, leftmargin=1em]
        \item \textit{Single-choice}: respond with exactly one option letter followed by a period, e.g., \texttt{A.}
        \item \textit{Multi-choice}: respond with all valid option letters separated by commas, followed by a period, e.g., \texttt{A, C.}
        \item \textit{Ordered-sequence}: respond with the sequence of phase letters in exact chronological order, merging consecutive identical phases while preserving non-consecutive recurrences, e.g., \texttt{A, B, A, D.}
    \end{itemize}
\end{enumerate}

An illustrative system prompt for a single-sequence, HBB-grounded memory task~(Object Appear Memory) is shown below to demonstrate how these five components are composed in practice.

\begin{systempromptbox}[Illustrative System Prompt --- Object Appear Memory (HBB-Grounded)]
\small
You are an expert in remote sensing image analysis, specializing in temporal land cover analysis.

\medskip
You will receive a temporal sequence of remote sensing images. Each image is preceded by its year label in the format:\\
\texttt{YYYY: [image]}\\
Images are arranged in chronological order from earliest to latest.

\medskip
Objects are located using Horizontal Bounding Boxes (HBB Box) in the format $\langle$x\_min, y\_min, x\_max, y\_max$\rangle$, where:
\begin{itemize}[nosep, leftmargin=1.5em]
    \item x\_min: left boundary coordinate
    \item y\_min: top boundary coordinate
    \item x\_max: right boundary coordinate
    \item y\_max: bottom boundary coordinate
\end{itemize}
Note: All coordinate values are normalized to the range [0, 999], where (0, 0) represents the top-left corner and (999, 999) represents the bottom-right corner of the image.

\medskip
Land cover is classified into five categories:
\begin{enumerate}[nosep, leftmargin=1.5em]
    \item \textbf{vegetation} --- areas covered by plants, grass, trees, or crops
    \item \textbf{buildings} --- man-made structures such as houses, factories, or commercial buildings
    \item \textbf{non-vegetated surface} --- bare soil, roads, parking lots, or other exposed ground
    \item \textbf{water} --- rivers, lakes, ponds, or other water bodies
    \item \textbf{playground} --- sports fields, recreational areas, or athletic facilities
\end{enumerate}

\medskip
Your task: Given a land cover object identified by its type and HBB Box at a specific time point, determine WHEN the current continuous presence of this land cover type began by tracing backward through the temporal sequence. Starting from the given time point, look backward until you find the transition from a different land cover type to the current one --- that transition marks when this land cover type most recently emerged at the specified location.

\medskip
This is a single-choice question. Select the time interval that best represents when the current continuous presence of this land cover type began at the given location.

\medskip
IMPORTANT: Respond with ONLY the option letter followed by a period (e.g., ``A.'' or ``B.''). Do not provide explanations or reasoning.
\end{systempromptbox}

\subsubsection{User Prompt Organization}\label{sec:user_prompt}

The user prompt is dynamically constructed and consists of two layers: an interleaved time--image context followed by a textual QA suffix.

For single-sequence tasks, the user prompt is organized as:

\begin{quote}\small\ttfamily
YYYY\textsubscript{1}: [IMAGE]\\
YYYY\textsubscript{2}: [IMAGE]\\
\ldots\\
YYYY\textsubscript{T}: [IMAGE]\\[4pt]
\textrm{\textit{$\langle$Question text$\rangle$}}\\
Choose from: \textrm{\textit{$\langle$options\_str$\rangle$}}
\end{quote}

\noindent where images are ordered chronologically from the earliest to the latest observation. For cross-sequence tasks, the prompt prepends region labels and preserves each region's own chronological order without merging the two sequences into a single global timeline:

\begin{quote}\small\ttfamily
Region\#1:\\
YYYY\textsubscript{1}: [IMAGE]\\
\ldots\\
YYYY\textsubscript{m}: [IMAGE]\\[4pt]
Region\#2:\\
YYYY\textsubscript{1}: [IMAGE]\\
\ldots\\
YYYY\textsubscript{n}: [IMAGE]\\[4pt]
\textrm{\textit{$\langle$Question text$\rangle$}}\\
Choose from: \textrm{\textit{$\langle$options\_str$\rangle$}}
\end{quote}

\subsubsection{Output Parsing and Evaluation Metrics}\label{sec:metrics}

\paragraph{Answer extraction.}
The benchmark does not directly compare raw model generations. Instead, we apply a multi-stage regex-based parser to extract option letters from the model output. The parser first attempts to match structured patterns such as comma-separated letter lists~(e.g., \texttt{A, C.}), then falls back to keyword-anchored patterns~(e.g., ``\texttt{The answer is B}'', ``\texttt{option A}''), and finally attempts to extract isolated standalone letters for very short outputs. All matched letters are converted to uppercase. For single-choice and multi-choice tasks, the extracted letters form an unordered set~$P$; for ordered-sequence tasks, the extraction preserves the original order to produce a list~$P = [p_1, \dots, p_m]$. If no valid letter can be extracted, the parser returns an empty set~(or empty list), and the sample is scored as incorrect.

\paragraph{Accuracy computation.}
We define accuracy separately for the three answer formats. For \textbf{single-choice} and \textbf{multi-choice} tasks, a prediction is correct only under exact match, i.e., the parsed set equals the gold set~($P = G$). For \textbf{ordered-sequence} tasks~(used exclusively for Object History Memory), a prediction is correct only if the parsed list matches the gold list in both content and order~($P = G$). The headline overall accuracy~(OA) aggregates exact-match correctness across all three formats:
\begin{equation}\label{eq:oa}
\mathrm{OA} = \frac{\text{exact\_match}_{\mathrm{single}} + \text{exact\_match}_{\mathrm{multi}} + \text{exact\_match}_{\mathrm{os}}}{N_{\mathrm{total}}}.
\end{equation}
Category-level accuracy~(\textbf{AVG}) is computed analogously over the subset of samples belonging to each of the four cognitive levels.

\subsection{Training Settings}\label{sec:train_settings}

GeoChrono is built upon Qwen3-VL-4B-Instruct~\cite{bai2025qwen3} and trained on ChronoInstruct for one epoch. The base model employs a patch size of $16\times16$; both training and inference use an input resolution of $1024\times1024$. The Vision Encoder produces per-frame visual features of length $4{,}096$, which the Vision-Language Projector then produces via $2\times2$ spatial pooling to a per-frame feature length of $S=1{,}024$ with an embedding dimension of $D=2{,}560$.

\paragraph{TempEnc configuration.}
TempEnc adopts a single-layer attention architecture with $H=16$ heads. In the Hybrid Temporal Attention, the first $H/2=8$ heads operate with bidirectional self-attention to capture global temporal context, while the remaining $8$ heads employ causal self-attention to model directional evolution. For the Semantic Focusing cross-attention, the text embeddings~$E_{\mathrm{txt}}$ serving as keys and values are derived from the concatenation of the system prompt and user prompt embeddings produced by the LLM's embedding layer.

\paragraph{C2FComp configuration.}
C2FComp partitions each frame's spatial tokens into non-overlapping blocks of size $b\times b = 2\times2$ during the Spatial Block Partitioning stage, yielding $S_c = S / b^2 = 256$ compressed blocks per frame. In the Prompt-Aware Scoring stage, the cross-attention module reuses the same text embeddings~$E_{\mathrm{txt}}$ as TempEnc.

\paragraph{Training strategy.}
During training, the Vision Encoder and Vision-Language Projector are kept frozen to preserve the pre-trained visual representations. The LLM backbone is fine-tuned via LoRA~\cite{hu2022lora}, while both TempEnc and C2FComp are randomly initialized and fully fine-tuned. The complete set of training hyperparameters is summarized in Table~\ref{tab:train_config}.

\begin{table}[htb]
\centering
\caption{Training configuration of GeoChrono.}
\label{tab:train_config}
\small
\setlength{\tabcolsep}{10pt}
\renewcommand{\arraystretch}{1.15}
\begin{tabular}{ll}
\toprule
\textbf{Configuration} & \textbf{Details} \\
\midrule
\multicolumn{2}{c}{\textit{Model Configuration}} \\
Base Model & Qwen3-VL-4B-Instruct \\
Input Resolution & $1024 \times 1024$ \\
LoRA Rank & 32 \\
LoRA Alpha & 64 \\
LoRA Dropout & 0.0 \\
\midrule
\multicolumn{2}{c}{\textit{Training Configuration}} \\
Hardware & 4 $\times$ NVIDIA H100 80GB \\
Global Batch Size & 64 \\
Training Duration & 1 epoch \\
\midrule
\multicolumn{2}{c}{\textit{Optimizer Configuration}} \\
Optimizer & AdamW \\
LR (LLM / LoRA) & $1 \times 10^{-4}$ \\
LR (TempEnc \& C2FComp) & $5 \times 10^{-4}$ \\
LR Scheduler & Cosine \\
Warm-up Ratio & 0.03 \\
\bottomrule
\end{tabular}
\end{table}

\begin{table}[htb]
\centering
\caption{Quantitative evaluation results on ChronoBench. We compare human expert performance against commercial MLLMs, open-source general MLLMs, remote sensing domain models, and our GeoChrono. All values are reported in percentage (\%).}
\label{tab:full_results}
\setlength{\tabcolsep}{3pt}
\renewcommand{\arraystretch}{1.10}
\resizebox{\textwidth}{!}{
\begin{tabular}{
l
c c c!{\vrule width 1pt}
c c c c c c!{\vrule width 1pt}
c c c c c c c!{\vrule width 1pt}
c c c c c c c!{\vrule width 1pt}
c
}
\toprule
\multirow{3}{*}{\textbf{Method}}
& \multicolumn{3}{>{\columncolor{LCPColor}}c!{\vrule width 1pt}}{\textbf{Land Cover Perception}}
& \multicolumn{6}{>{\columncolor{TAColor}}c!{\vrule width 1pt}}{\textbf{Temporal Recognition}}
& \multicolumn{7}{>{\columncolor{LTMColor}}c!{\vrule width 1pt}}{\textbf{Long-Term Memory}}
& \multicolumn{7}{>{\columncolor{STRColor}}c!{\vrule width 1pt}}{\textbf{Spatio-Temporal Reasoning}}
& \multirow{3}{*}{\textbf{OA}} \\
\cmidrule(lr){2-4}
\cmidrule(lr){5-10}
\cmidrule(lr){11-17}
\cmidrule(lr){18-24}
& \multicolumn{2}{c}{\textbf{OP}}
& \multirow{2}{*}{\textbf{AVG}}
& \multirow{2}{*}{\textbf{BCAC}}
& \multirow{2}{*}{\textbf{BCC}}
& \multirow{2}{*}{\textbf{LCC}}
& \multicolumn{2}{c}{\textbf{BOC}}
& \multirow{2}{*}{\textbf{AVG}}
& \multicolumn{2}{c}{\textbf{OAM}}
& \multicolumn{2}{c}{\textbf{OCM}}
& \multicolumn{2}{c}{\textbf{OHM}}
& \multirow{2}{*}{\textbf{AVG}}
& \multirow{2}{*}{\textbf{CCME}}
& \multirow{2}{*}{\textbf{RDC}}
& \multicolumn{2}{c}{\textbf{OCO}}
& \multicolumn{2}{c}{\textbf{CSOCO}}
& \multirow{2}{*}{\textbf{AVG}}
& \\
\cmidrule(lr){2-3}
\cmidrule(lr){8-9}
\cmidrule(lr){11-12}
\cmidrule(lr){13-14}
\cmidrule(lr){15-16}
\cmidrule(lr){20-21}
\cmidrule(lr){22-23}
& \textbf{Coord} & \textbf{Box}
&
&
&
&
& \textbf{Coord} & \textbf{Box}
&
& \textbf{Coord} & \textbf{Box}
& \textbf{Coord} & \textbf{Box}
& \textbf{Coord} & \textbf{Box}
&
&
&
& \textbf{Coord} & \textbf{Box}
& \textbf{Coord} & \textbf{Box}
&
& \\
\midrule
\textbf{Human}
& 98.52 & 95.56 & 97.04
& 94.81 & 82.96 & 82.96
& 98.52 & 89.63
& 89.78
& 98.52 & 97.04
& 91.11 & 87.41
& 85.93 & 90.37
& 91.73
& 78.52 & 97.78
& 99.26 & 100.00
& 99.26 & 98.52
& 95.56 & 92.28 \\
\midrule

\rowcolor{GroupGray}
\multicolumn{25}{l}{\textbf{Commercial models}} \\
\specialrule{1pt}{0pt}{0pt}
Gemini-3-Flash\cite{team2023gemini}
& 65.06 & 65.96 & 65.51
& 82.00 & 50.53 & 52.45
& 55.85 & 56.16
& 61.38
& 56.44 & 58.44
& 46.68 & 47.75
& 25.00 & 21.36
& 47.52
& 37.79 & \bestSTR{63.64}
& \bestSTR{68.05} & 78.01
& 54.36 & 66.67
& 59.89 & 57.48 \\
GPT-5.4\cite{openai2025gpt41}
& 47.07 & 40.00 & 43.53
& 80.00 & 50.76 & 53.87
& 66.13 & 60.00
& 67.57
& 47.69 & 41.52
& 35.36 & 46.09
& 26.14 & 20.00
& 39.21
& 30.73 & 30.51
& 53.83 & 70.60
& 52.43 & 69.12
& 50.42 & 56.29 \\
Seed-1.6-Vision\cite{bytedance2026seed16}
& 50.98 & 73.23 & 62.11
& 81.91 & 46.27 & 53.73
& 47.08 & 58.47
& 63.87
& 37.08 & 47.05
& 37.22 & 41.85
& 20.00 & 18.41
& 36.86
& 31.22 & 38.98
& 56.52 & 73.29
& 55.83 & 79.41
& 53.74 & 55.48 \\
\midrule

\rowcolor{GroupGray}
\multicolumn{25}{l}{\textbf{Open-source models}} \\
\specialrule{1pt}{0pt}{0pt}
InternVL-3.5-4B\cite{wang2025internvl3}
& 45.41 & 58.65 & 52.03
& 54.83 & 32.57 & 27.95
& 24.51 & 24.20
& 38.19
& 22.03 & 21.49
& 6.09 & 10.20
& 1.36 & 0.91
& 13.33
& 22.76 & 33.33
& 54.66 & 49.28
& 50.49 & 52.45
& 42.07 & 33.25 \\
InternVL-3.5-8B\cite{wang2025internvl3}
& 33.08 & 45.56 & 39.32
& 56.29 & 28.31 & 28.49
& 35.60 & 39.35
& 43.21
& 25.11 & 24.30
& 13.91 & 16.95
& 1.60 & 3.41
& 16.93
& 30.40 & 39.55
& 51.35 & 53.21
& 57.77 & 47.55
& 45.11 & 36.32 \\
InternVL-3.5-14B\cite{wang2025internvl3}
& 40.90 & 55.94 & 48.42
& 64.15 & 33.41 & 30.66
& 36.43 & 30.08
& 45.67
& 22.39 & 21.76
& 15.76 & 19.07
& 0.45 & 0.91
& 16.45
& 21.95 & 28.25
& 52.38 & 53.83
& 48.06 & 49.51
& 41.42 & 37.76 \\
Qwen3-VL-4B\cite{bai2025qwen3}
& 35.19 & 41.35 & 38.27
& 66.94 & 31.13 & 36.77
& 23.96 & 16.73
& 42.07
& 22.76 & 18.95
& 18.94 & 22.52
& 2.95 & 4.55
& 17.54
& 21.30 & 23.16
& 47.83 & 47.83
& 54.85 & 53.92
& 39.53 & 35.10 \\
Qwen3-VL-8B\cite{bai2025qwen3}
& 39.40 & 41.96 & 40.68
& 69.21 & 29.15 & 43.15
& 31.98 & 29.26
& 47.12
& 29.74 & 28.56
& 17.48 & 20.00
& 2.95 & 0.91
& 20.52
& 23.74 & 28.81
& 54.66 & 53.00
& 55.83 & 47.06
& 42.80 & 39.19 \\
Qwen3-VL-32B\cite{bai2025qwen3}
& 42.86 & 44.66 & 43.76
& 75.42 & 44.29 & 47.90
& 45.68 & 46.16
& 57.88
& 29.92 & 30.73
& 25.03 & 27.95
& 3.86 & 4.55
& 24.06
& 28.78 & 32.77
& 51.76 & 59.83
& 59.71 & 62.25
& 47.23 & 46.73 \\
\midrule

\rowcolor{GroupGray}
\multicolumn{25}{l}{\textbf{Remote sensing domain models}} \\
\specialrule{1pt}{0pt}{0pt}
TEOChat-7B\cite{irvin2024teochat}
& 32.63 & 36.99 & 34.81
& 48.85 & 14.23 & 4.21
& 17.27 & 24.25
& 30.07
& 24.75 & 27.83
& 4.11 & 6.62
& 1.59 & 1.14
& 14.64
& 17.40 & 27.68
& 50.93 & 53.21
& 9.22 & 22.06
& 33.35 & 26.82 \\
EarthDial-4B\cite{soni2025earthdial}
& 36.99 & 40.30 & 38.65
& 54.83 & 18.04 & 4.34
& 17.72 & 24.09
& 33.08
& 26.20 & 26.20
& 17.48 & 17.75
& 0.68 & 1.36
& 18.56
& 15.61 & 22.03
& 48.44 & 46.38
& 41.75 & 52.45
& 36.25 & 30.11 \\
DVLChat-4B\cite{xuan2025dynamicvl}
& 48.57 & 46.92 & 47.74
& 85.90 & 31.81 & 37.18
& 34.21 & 30.46
& 54.48
& 29.56 & 26.93
& 24.37 & 29.93
& 2.50 & 2.05
& 22.91
& 17.24 & 25.42
& 50.72 & 51.76
& 58.74 & 55.39
& 40.59 & 44.07 \\
\midrule

\rowcolor{GroupGray}
\multicolumn{25}{l}{\textbf{Ours}} \\
\specialrule{1pt}{0pt}{0pt}
GeoChrono
& \bestLCP{83.91} & \bestLCP{93.38} & \bestLCP{88.65}
& \bestTA{93.66} & \bestTA{65.91} & \bestTA{76.66}
& \bestTA{77.83} & \bestTA{80.27}
& \bestTA{83.03}
& \bestLTM{72.71} & \bestLTM{87.04}
& \bestLTM{68.61} & \bestLTM{75.10}
& \bestLTM{31.59} & \bestLTM{32.73}
& \bestLTM{68.10}
& \bestSTR{62.60} & 61.58
& 61.49 & \bestSTR{92.55}
& \bestSTR{75.24} & \bestSTR{92.16}
& \bestSTR{72.92} & \bestOA{78.34} \\

\bottomrule
\end{tabular}
}
\end{table}

\section{Experimental Results}\label{sec:exp_results}

\subsection{Benchmark Result Analysis}\label{sec:bench_analysis}

\subsubsection{Impact of Spatial Grounding Modality}\label{sec:grounding_analysis}

Seven of the twelve sub-tasks in ChronoBench~(i.e., OP, BOC, OAM, OCM, OHM, OCO, and CSOCO) provide two spatial grounding variants: \textit{HBB Box}, which localizes the target via a horizontal bounding box in normalized image coordinates $[0,999]$, and \textit{Geo Coordinate}, which specifies the coordinate reference system, the image corner coordinates, and an interior point of the target object. HBB Box is the most prevalent grounding modality in existing MLLM benchmarks. Geo-coordinate grounding, by contrast, requires the model to first infer the target point's relative position within the image extent defined by the corner coordinates, and then extrapolate the target's identity from the local semantic context around that point, posing a considerably greater challenge for spatial reasoning.

Table~\ref{tab:full_results} confirms this asymmetry across the board. For the vast majority of models and tasks, the HBB-Box variant yields higher accuracy than its coordinate-grounded counterpart. The gap is particularly pronounced on cognitively demanding tasks: for example, on OAM, Gemini-3-Flash scores 58.44\% with HBB Box versus 56.44\% with coordinates. In the few instances where coordinate-grounded accuracy exceeds that of HBB Box---such as certain models on OHM---the absolute performance is extremely low~(e.g., below 5\%), indicating that the model's capacity for the underlying temporal reasoning task is itself near chance level, and the observed reversal is attributable to stochastic fluctuation rather than genuine coordinate-grounding advantage.

Based on this observation, subsequent fine-grained analyses in this section focus exclusively on the HBB-Box-grounded subset of each task, thereby isolating the model's temporal understanding ability from the confounding factor of spatial grounding difficulty.

\subsubsection{Land-Cover Perception Confusion Analysis}\label{sec:perception_confusion}

\begin{figure}[tb]
    \centering
    \includegraphics[width=0.8\linewidth]{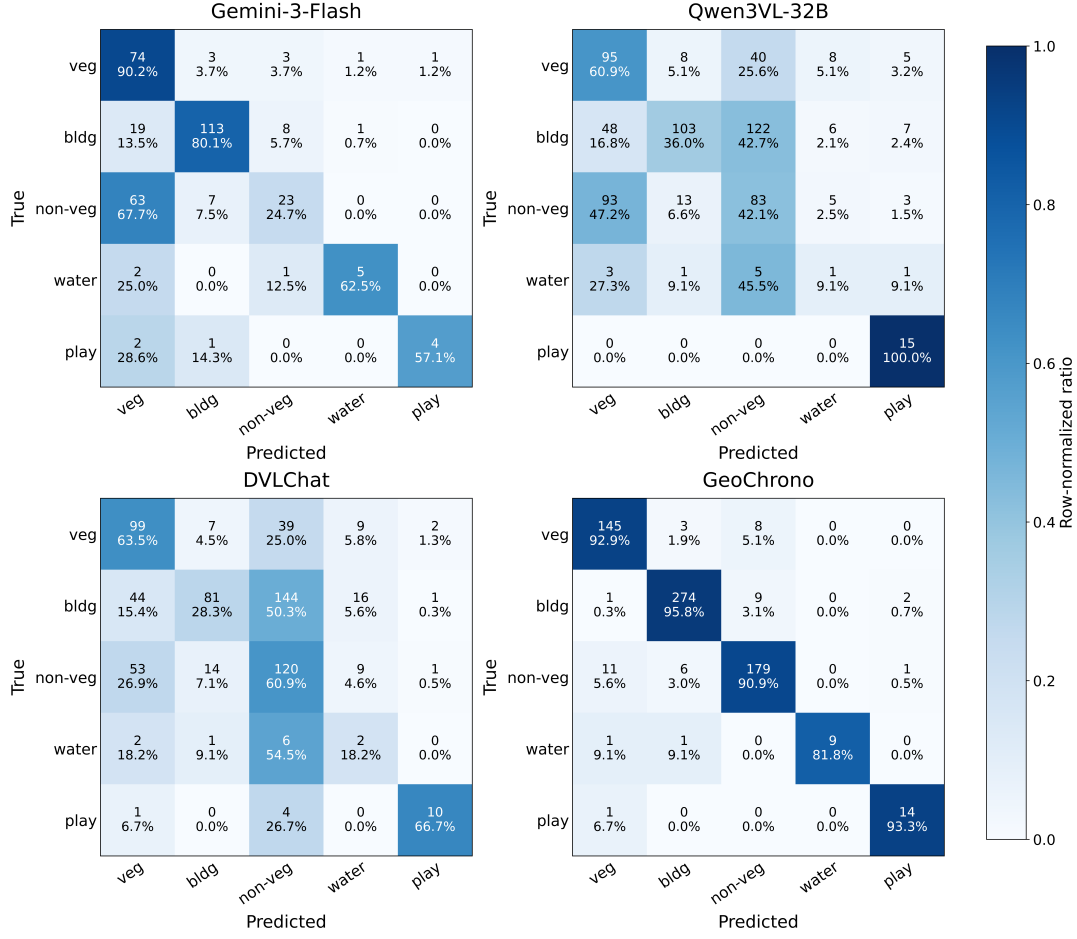}
    \caption{Row-normalized confusion matrices for the HBB-grounded Object Perception~(OP) task. Each cell shows the count and percentage of samples with a given true class~(row) predicted as the indicated class~(column). Four representative models are compared: Gemini-3-Flash, Qwen3-VL-32B, DVLChat, and GeoChrono.}
    \label{fig:confusion_matrix}
\end{figure}

Accurate land-cover perception underpins all higher-level temporal reasoning tasks in ChronoBench: a model that cannot reliably identify the current land-cover type at a given location will inevitably propagate errors into change detection, memory retrieval, and spatio-temporal reasoning. To examine this foundational capability in detail, we select one representative model from each of the three evaluated categories, i.e., Gemini-3-Flash~(commercial models), Qwen3-VL-32B~(open-source general MLLMs), and DVLChat~(RS domain MLLMs), together with GeoChrono, and compute per-class confusion matrices on the HBB-grounded Object Perception~(OP) task. The results are visualized in Figure~\ref{fig:confusion_matrix}.

A dominant source of confusion shared by Gemini-3-Flash, Qwen3-VL-32B, and DVLChat is the mutual misclassification between \textit{vegetation} and \textit{non-vegetated surface}. For instance, Qwen3-VL-32B misclassifies 25.6\% of vegetation samples as non-vegetated surface and 47.2\% of non-vegetated surface samples as vegetation; DVLChat similarly confuses these two classes at rates of 25.0\% and 26.9\%, respectively. This confusion is largely attributable to the seasonal variability inherent in NAIP imagery: because acquisitions span different months across years, vegetation cover can undergo substantial phenological changes. These changes appear brown and dormant in winter or early spring, closely resembling bare soil or exposed ground. This systematic challenge serves as a natural stress test for a model's seasonal robustness in land-cover classification. In contrast, GeoChrono achieves 92.9\% accuracy on vegetation and 90.9\% on non-vegetated surface, with cross-confusion rates below 6\%, demonstrating strong resilience to seasonal appearance variation.

A second notable error pattern involves \textit{buildings} being misclassified as \textit{non-vegetated surface}. In Qwen3-VL-32B, 42.7\% of building samples are predicted as non-vegetated surface; DVLChat also exhibits a high confusion rate of 50.3\% on this pair. While buildings exhibit distinctive optical features, such as regular geometry and shadow patterns, they often occupy small spatial extents within wide-area remote sensing images, making precise localization within the HBB-grounded region particularly challenging. GeoChrono reduces this confusion to 3.1\%, which we attribute to TempEnc's spatial context aggregation mechanism that enriches each spatial token with local neighborhood information before temporal modeling, thereby strengthening fine-grained feature discrimination even for small-footprint objects.

Overall, GeoChrono achieves diagonal-dominant confusion matrices across all five land-cover classes, with per-class accuracy exceeding 81\% for every category and reaching above 90\% for four out of five classes. This robust perceptual foundation provides a reliable basis for the downstream temporal recognition, memory, and reasoning tasks evaluated in ChronoBench.

\subsubsection{Object History Memory: Accuracy under various Ground-Truth Sequence Length}\label{sec:ohm_length}

\begin{figure}[htb]
    \centering
    \includegraphics[width=0.8\linewidth]{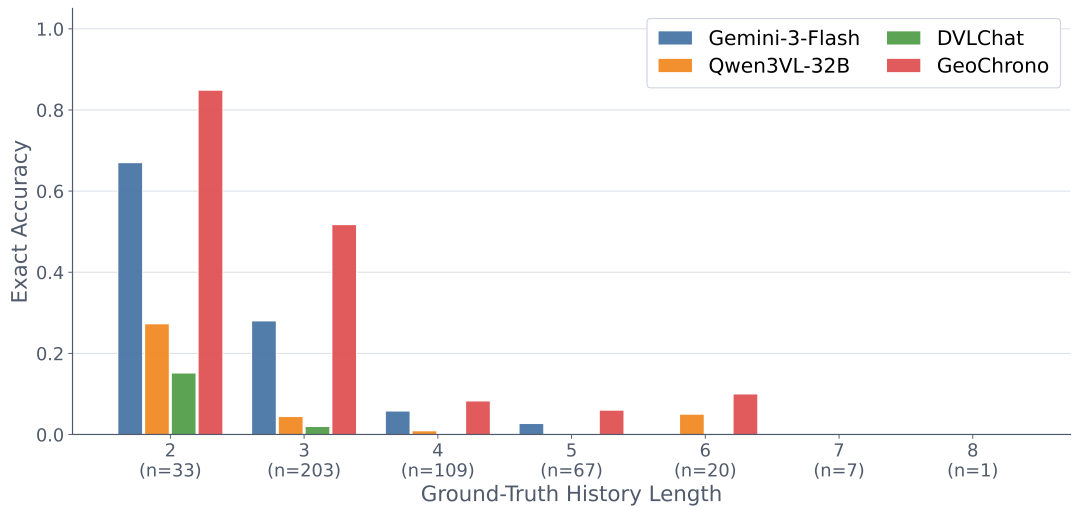}
    \caption{Exact-match accuracy on HBB-grounded Object History Memory~(OHM) stratified by ground-truth evolution chain length. The number of samples at each length is indicated in parentheses. Four representative models are compared.}
    \label{fig:ohm_length}
\end{figure}

Object History Memory~(OHM) requires the model to reconstruct the complete chronological chain of land-cover phases at a given location, making it the most demanding sub-task in Long-Term Memory tasks. As shown in Table~\ref{tab:full_results}, even the best-performing baseline, Gemini-3-Flash, achieves only 21.36\% exact-match accuracy on HBB-grounded OHM, while open-source and RS domain models largely remain below 5\%. To gain deeper insight into this failure mode, we stratify the HBB-grounded OHM samples by the length of their ground-truth evolution chain and compare the exact-match accuracy of four representative models in Figure~\ref{fig:ohm_length}.

The results reveal a clear inverse relationship between chain length and accuracy across all models. For the shortest chains~(length\,=\,2), identifying a single land-cover transition is relatively straightforward, and GeoChrono substantially outperforms all baselines. At length\,=\,3, which constitutes the largest sample group~($n$\,=\,203), accuracy drops sharply for all baselines. Qwen3-VL-32B and DVLChat fall to near zero, while Gemini-3-Flash also declines considerably, yet GeoChrono retains a clear lead. Beyond length\,=\,3, baseline models essentially collapse, with Qwen3-VL-32B and DVLChat unable to correctly reconstruct evolution chains and Gemini-3-Flash producing only sporadic correct predictions. GeoChrono, while also declining, remains the only model that achieves non-trivial accuracy at lengths 4 through 6, demonstrating its capacity to partially recover longer evolution histories. Note that samples with chain lengths of 7 and 8 are extremely scarce~($n$\,=\,7 and $n$\,=\,1), making accuracy estimates at these lengths unreliable.

This length-dependent degradation exposes a fundamental limitation: existing MLLMs lack the capacity to track cumulative state transitions across extended temporal sequences. The consistent advantage of GeoChrono across all chain lengths validates the design of TempEnc, whose per-location trajectory modeling explicitly constructs a dedicated temporal signal for each spatial position, enabling the model to trace sequential land-cover transitions rather than relying on holistic frame-level comparisons. Nevertheless, the rapid decline in accuracy beyond length\,=\,3 also indicates that long-chain history reconstruction remains an open challenge, warranting further research into more expressive temporal memory architectures.

\subsection{Computational Efficiency Analysis}\label{sec:efficiency}

To quantify the computational overhead introduced by each architectural component, we conduct controlled inference profiling on a single NVIDIA H100 80GB GPU. The profiling uses a simulated input of $T=7$ images at $1024\times1024$ resolution with $32$ output tokens, deployed with Flash-Attention-2 and \texttt{bfloat16} precision. Each configuration is measured over 10 runs following 1 warm-up iteration. We take the direct fine-tuned model~(without TempEnc or C2FComp) as the baseline and progressively add components. All results are summarized in Table~\ref{tab:efficiency}.

\begin{table}[tb]
\centering
\caption{Computational efficiency profiling of GeoChrono's architectural components. ``DFT'' denotes the baseline without TempEnc or C2FComp. Selection ratio refers to the fraction of spatial blocks retained at full resolution by C2FComp~($K/S_c$). Peak memory reports the maximum GPU allocation during inference. Relative changes are computed against the LoRA baseline.}
\label{tab:efficiency}
\small
\setlength{\tabcolsep}{4pt}
\renewcommand{\arraystretch}{1.12}
\resizebox{0.8\columnwidth}{!}{
\begin{tabular}{lccccc}
\toprule
\multirow{2}{*}{\textbf{Configuration}} & \textbf{Selection} & \textbf{Peak Mem.} & \textbf{FLOPs} & \textbf{Visual Tokens} & \textbf{OA} \\
 & \textbf{Ratio} & \textbf{(GB)} & \textbf{(TFLOPs)} & \textbf{(per sample)} & \textbf{(\%)} \\
\midrule
DFT~(baseline)            & ---   & 12.04 & 85.07 & 7,168  & 72.52 \\
+ TempEnc                  & ---   & 12.14\,{\scriptsize(+0.8\%)} & 85.65\,{\scriptsize(+0.7\%)} & 7,168\,{\scriptsize(+0.0\%)} & 78.34 \\
\midrule
+ TempEnc + C2FComp        & 1/2   & 11.04\,{\scriptsize(-8.3\%)} & 52.05\,{\scriptsize(-38.8\%)} & 4,480\,{\scriptsize(-37.5\%)} & 74.58 \\
+ TempEnc + C2FComp        & 1/4   & 10.46\,{\scriptsize(-13.1\%)} & 37.22\,{\scriptsize(-56.2\%)} & 3,136\,{\scriptsize(-56.3\%)} & 74.11 \\
+ TempEnc + C2FComp        & 1/16  & 10.30\,{\scriptsize(-14.5\%)} & 26.98\,{\scriptsize(-68.3\%)} & 2,128\,{\scriptsize(-70.3\%)} & 72.60 \\
+ TempEnc + C2FComp        & 1/64  & 10.30\,{\scriptsize(-14.5\%)} & 24.53\,{\scriptsize(-71.2\%)} & 1,876\,{\scriptsize(-73.8\%)} & 71.56 \\
\bottomrule
\end{tabular}
}
\end{table}

\paragraph{TempEnc introduces negligible overhead.}
Adding TempEnc to the DFT baseline increases peak GPU memory by only 0.8\%~(12.04$\to$12.14\,GB) and total FLOPs by 0.7\%~(85.07$\to$85.65\,TFLOPs), while the number of visual tokens remains unchanged at 7,168. Despite this marginal computational cost, TempEnc delivers a substantial performance improvement of 5.82\% OA~(72.52\%$\to$78.34\%), confirming that TempEnc operates efficiently within the existing visual token budget without inflating the sequence forwarded to the LLM.

\paragraph{C2FComp enables flexible efficiency--performance trade-offs.}
Once C2FComp is activated, all three efficiency metrics (i.e., peak memory, FLOPs, and visual token count) decrease monotonically as the selection ratio decreases. At a selection ratio of 1/4~($K=64$), C2FComp reduces visual tokens by 56.3\%~(7,168$\to$3,136) and total FLOPs by 56.2\%, while retaining 94.6\% of the full model's performance~(74.11\% vs.\ 78.34\% OA). Peak memory also drops by 13.1\%. This operating point provides an effective balance between computational efficiency and performance. More aggressive compression at 1/16 and 1/64 further reduces FLOPs by 68.3\% and 71.2\%, respectively, while retaining performance above 91\%, at the cost of diminishing model performance with each additional compression. Notably, peak memory saturates beyond 1/16~(10.30\,GB for both 1/16 and 1/64), as the remaining overhead is dominated by fixed costs~(vision encoder, LLM weights) that are independent of the visual token count.

\begin{figure}[!htbp]
    \centering
    \includegraphics[width=1\linewidth]{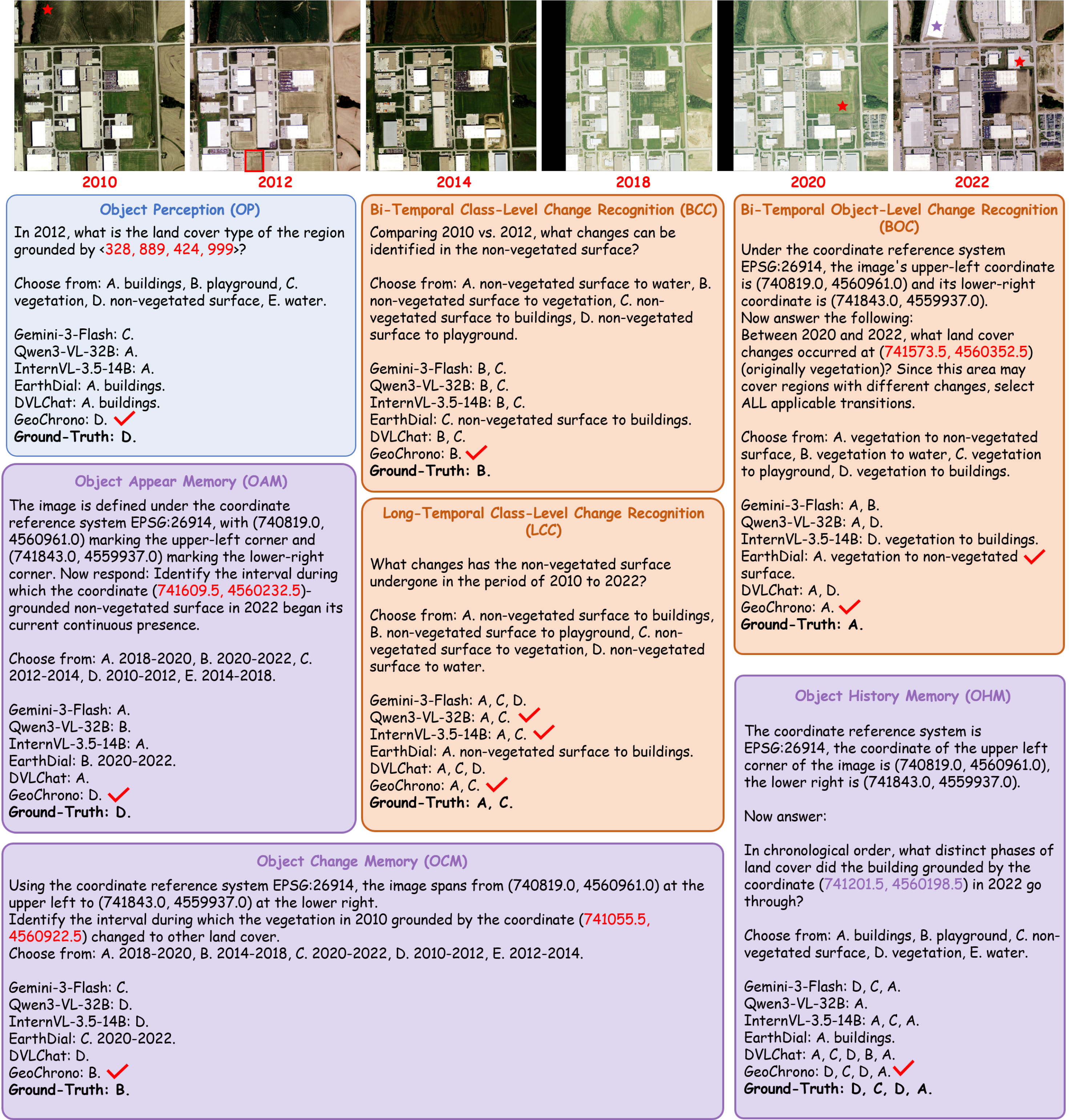}
    \caption{Qualitative results on Land Cover Perception~(OP), Temporal Recognition~(BCC, LCC, BOC), and Long-Term Memory~(OAM, OCM, OHM) sub-tasks. Correct predictions are marked with \textcolor{red}{\checkmark}. GeoChrono is the only model that correctly answers all seven tasks.}
    \label{fig:qual_1}
\end{figure}

\begin{figure}[!htbp]
    \centering
    \includegraphics[width=1\linewidth]{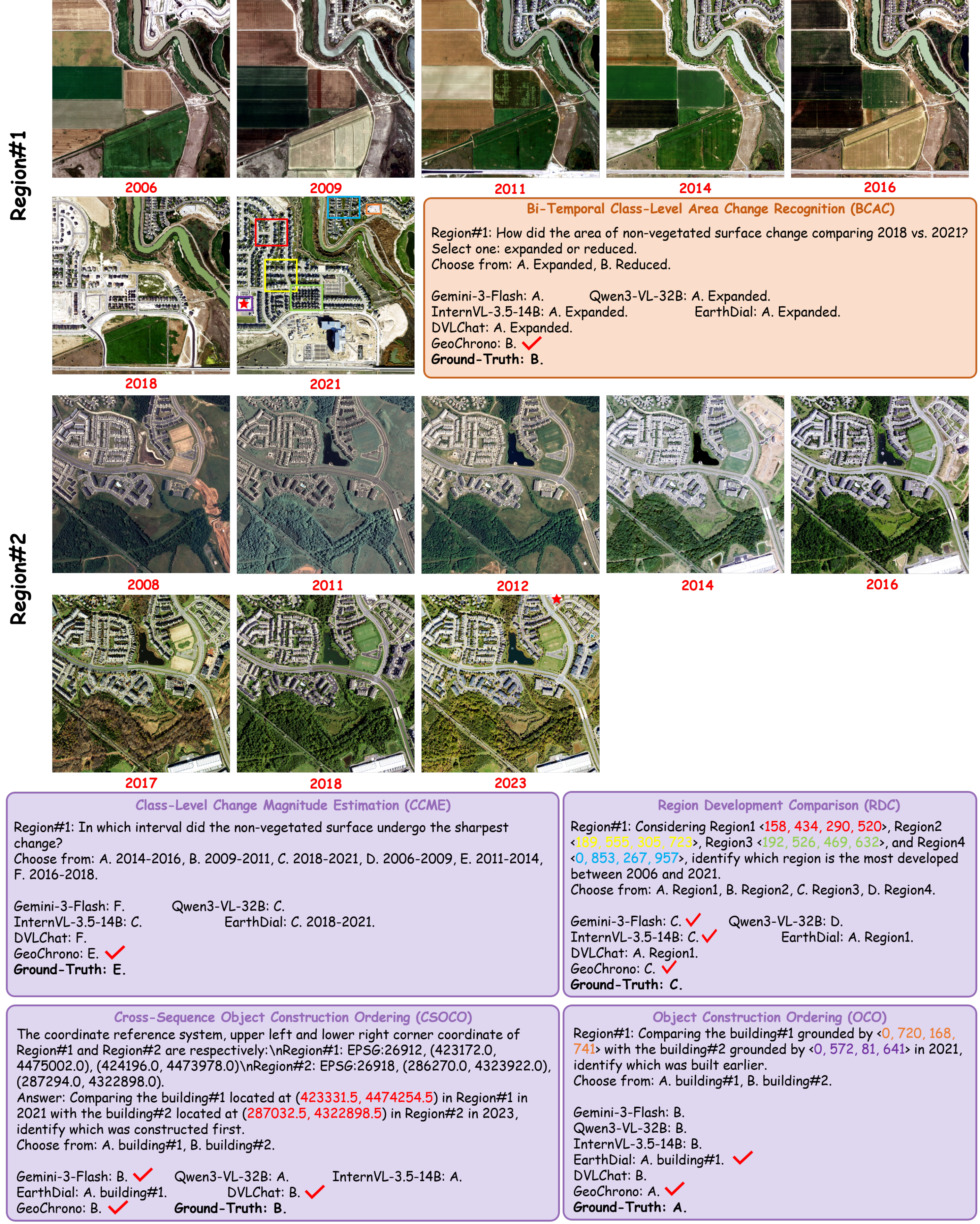}
    \caption{Qualitative results on Temporal Recognition~(BCAC) and Spatio-Temporal Reasoning~(CCME, RDC, OCO, CSOCO) sub-tasks. Correct predictions are marked with \textcolor{red}{\checkmark}. Except for CSOCO, all tasks take only the Region\#1 sequence as input; the ``Region\#1'' prefix shown before each prompt is added for readability and does not appear in the actual input.}
    \label{fig:qual_2}
\end{figure}

\subsection{Qualitative Results}\label{sec:qualitative}

Figures~\ref{fig:qual_1} and~\ref{fig:qual_2} present qualitative visualizations of GeoChrono alongside five representative baselines~(i.e., Gemini-3-Flash, Qwen3-VL-32B, InternVL-3.5-14B, EarthDial, and DVLChat) across all 12 sub-tasks of ChronoBench.

\paragraph{Land Cover Perception.}
In the OP example~(Figure~\ref{fig:qual_1}), the query targets a small region within the image. GeoChrono correctly identifies the land-cover type as \textit{non-vegetated surface}, while Gemini-3-Flash misclassifies it as \textit{vegetation}. The remaining four baselines all predict \textit{buildings}, a class with visually salient features in remote sensing imagery. This suggests that when the target occupies a small spatial extent, existing models tend to default to conspicuous categories rather than accurately localizing and interpreting the queried region.

\paragraph{Temporal Recognition.}
The BCC, LCC, and BOC examples in Figure~\ref{fig:qual_1} are all multi-choice tasks requiring the model to identify \emph{all} applicable land-cover transitions. GeoChrono precisely selects the correct set of transitions in each case. In contrast, the baselines consistently over-predict by selecting additional incorrect options. This tendency indicates that these models do not sufficiently rely on visual evidence to ground their change detection; instead, they appear to be influenced by the multi-choice instruction in the system prompt, defaulting to an overly inclusive selection strategy.
The BCAC example in Figure~\ref{fig:qual_2} presents a further challenge: between 2018 and 2021, \textit{non-vegetated surface} simultaneously gains area from some classes and loses area to others, requiring fine-grained comparison of the net change direction. All baselines incorrectly predict ``Expanded'', whereas the ground truth is ``Reduced''. Only GeoChrono yields the correct answer, demonstrating its ability to discriminate subtle area-balance shifts in bi-temporal change recognition.

\paragraph{Long-Term Memory.}
The OAM and OCM examples in Figure~\ref{fig:qual_1} expose a systematic temporal anchoring bias in existing models. In the OAM case, the queried object is a \textit{non-vegetated surface} observed in 2022 whose current continuous presence began during the 2010--2012 interval~(ground truth: D). However, most baselines select options temporally proximate to the reference year 2022, such as 2018--2020~(A) or 2020--2022~(B), rather than tracing backward to the actual emergence point. A similar pattern appears in the OCM case, where baselines gravitate toward options near the reference year 2010 or simply select the most recent available interval. This positional bias favors temporally adjacent or boundary options, indicating that these models lack genuine temporal event retrieval capability and instead rely on heuristic shortcuts. GeoChrono correctly identifies the target interval in both cases, demonstrating its ability to trace land-cover state transitions across the full temporal span. In the OHM example, which requires reconstructing a four-phase evolution chain~(D$\to$C$\to$D$\to$A), GeoChrono is the only model that produces the exact chronological sequence.

\paragraph{Spatio-Temporal Reasoning.}
The CCME example in Figure~\ref{fig:qual_2} demands comparison of change magnitudes across all consecutive intervals to identify the period of most substantial transformation. All baselines select plausible but incorrect distractor intervals, whereas GeoChrono correctly identifies the answer. This demonstrates its ability to perform comparative temporal reasoning over the full sequence. In the RDC, OCO, and CSOCO examples, GeoChrono consistently produces correct predictions, further validating its robust cross-region and cross-sequence spatio-temporal reasoning capabilities.

\begin{figure}[!htbp]
    \centering
    \includegraphics[width=1\linewidth]{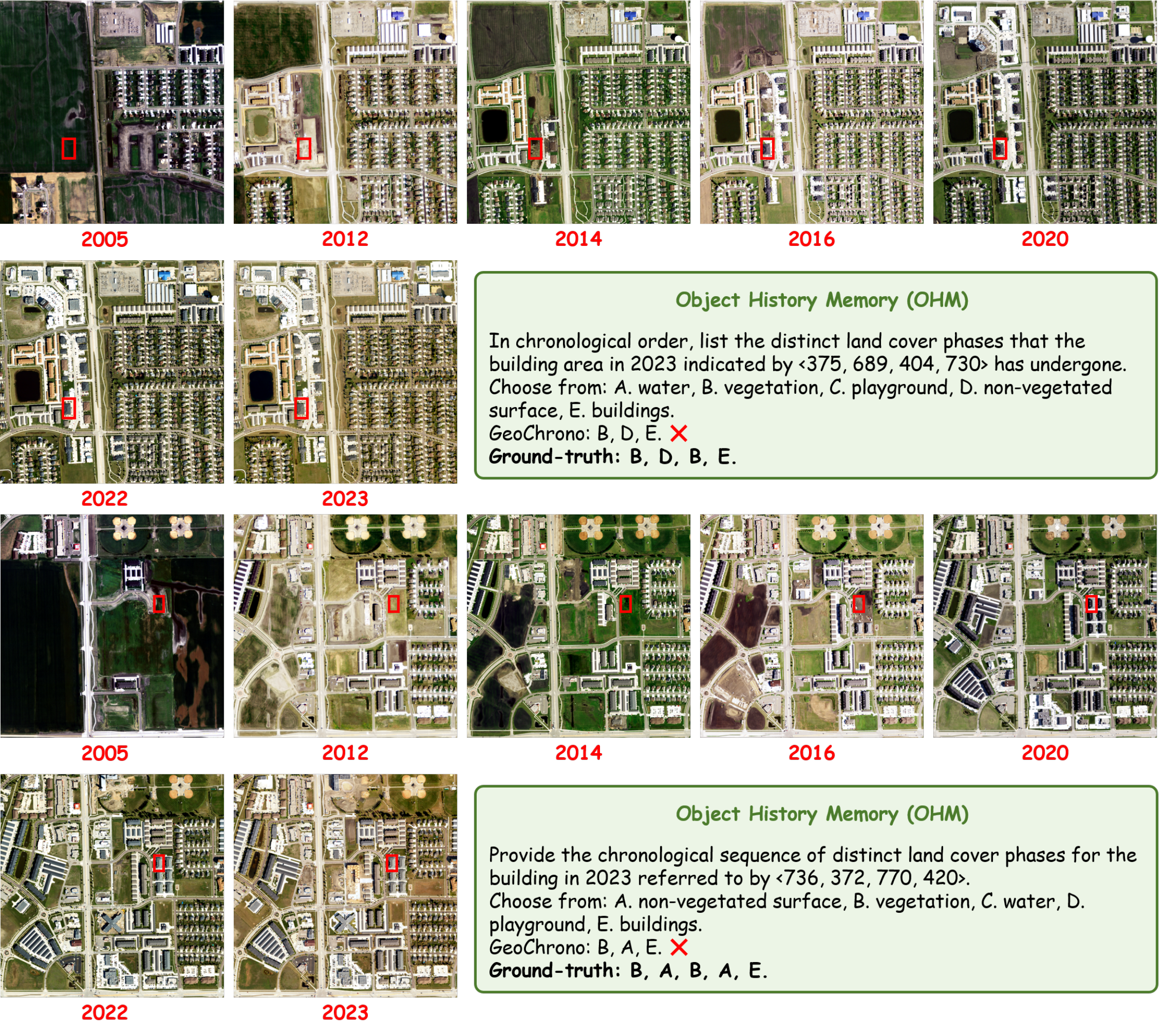}
    \caption{Two representative failure cases of GeoChrono on the OHM task. In both examples, the model omits intermediate land-cover phases in which the target location temporarily reverts to \textit{vegetation}, thereby producing shorter evolution chains. Ground-truth sequences and model predictions are shown; incorrect predictions are marked with \textcolor{red}{$\times$}.}
    \label{fig:failure_case}
\end{figure}

\subsection{Failure Case Study}\label{sec:failure_case}

Although GeoChrono substantially outperforms all other models on the Object History Memory~(OHM) task, a notable gap to human-level performance persists. We therefore conduct a targeted failure analysis on this task to identify systematic error patterns.

As shown in the accuracy--length analysis~(Section~\ref{sec:ohm_length}), model performance degrades substantially when the ground-truth evolution chain reaches length four or above. Figure~\ref{fig:failure_case} presents two representative failure cases that reveal a common error mode: \emph{phase omission}, where the model predicts a proper subsequence of the ground-truth chain by skipping intermediate phases.

In the first example, the target building undergoes a four-phase evolution: \textit{vegetation}~$\to$~\textit{non-vegetated surface}~$\to$~\textit{vegetation}~$\to$~\textit{buildings}. GeoChrono outputs B,~D,~E, correctly capturing the initial and final phases but missing the transient reversion to \textit{vegetation} between the two \textit{non-vegetated surface} intervals. The second example involves a five-phase chain: \textit{vegetation}~$\to$~\textit{non-vegetated surface}~$\to$~\textit{vegetation}~$\to$~\textit{non-vegetated surface}~$\to$~\textit{buildings}. Here, GeoChrono omits two intermediate phases, again failing to detect the periods during which the site temporarily returns to \textit{vegetation}.

Both failures share a common cause: the model struggles to recognize brief reversions to \textit{vegetation} that occur amid an overall development trajectory. In the \textit{vegetation} class, which encompasses both trees and shallow herbaceous cover, the visual appearance can closely resemble \textit{non-vegetated surface} depending on season and vegetation maturity. This subtle inter-class ambiguity between \textit{vegetation} and \textit{non-vegetated surface} poses significant challenges for recognition. These observations suggest that improving the robustness of per-frame land-cover discrimination, particularly for visually ambiguous transitional states, remains a critical direction for advancing long-horizon temporal memory in remote sensing MLLMs.

\section{Limitations and Future Directions}\label{sec:limitations}

While GeoChrono achieves state-of-the-art performance on ChronoBench and demonstrates strong temporal understanding capabilities in remote sensing, several limitations remain and point to promising future directions.
\textbf{i)}~Despite substantially outperforming all other leading MLLMs, GeoChrono still exhibits a notable gap relative to human experts~(78.34\% vs.\ 92.28\% OA).
\textbf{ii)}~The current framework operates exclusively on RGB imagery, yet NAIP acquisitions natively provide four-band data~(RGB + near-infrared). The near-infrared channel carries rich vegetation reflectance signatures that could significantly enhance land-cover discrimination, particularly for the visually ambiguous vegetation versus non-vegetated surface boundary identified in our confusion analysis~(Section~\ref{sec:perception_confusion}). Incorporating multi-spectral information into the visual encoder represents a straightforward yet potentially impactful extension.
\textbf{iii)}~The rich pixel-level change semantic masks supplied by the DVL-Suite~\cite{xuan2025dynamicvl} constitute a valuable yet underexploited supervisory signal in the current pipeline. Recent work has demonstrated that incorporating segmentation masks can effectively strengthen MLLM performance on visual question answering tasks~\cite{wang2026earthvl}. Leveraging such dense spatial annotations---either as auxiliary training supervision or as an additional input modality---to further enhance long-term remote sensing temporal understanding represents another promising avenue for future exploration.

\section{Contribution to the Community}\label{sec:contribution}

We hope the artifacts produced in this work can serve as a useful resource for the remote sensing and multimodal learning communities. Below, we discuss several aspects of its potential contribution, while acknowledging that much remains to be explored.

\paragraph{A structured evaluation lens for temporal understanding.}
Existing remote sensing benchmarks predominantly organize evaluation around application-specific scenarios~(e.g., change detection, damage assessment), making it difficult to pinpoint which underlying cognitive ability a model lacks. ChronoBench addresses this by decomposing long-term temporal understanding into four progressive cognitive levels (i.e., Land Cover Perception, Temporal Recognition, Long-Term Memory, and Spatio-Temporal Reasoning) and instantiating them as 12 concrete subtasks, with 17,689 validated QA pairs. We believe this structured taxonomy can offer the community a more diagnostic perspective: rather than reporting a single aggregate score, researchers can identify specific competency gaps and direct modeling efforts accordingly. We hope this evaluation paradigm can extend beyond the current land-cover setting and inspire analogous competency-based benchmarks for other temporal understanding domains.

\paragraph{Scalable and reproducible data construction.}
A recurring bottleneck in remote sensing research is the cost of high-quality temporal annotation. Our fully rule-based construction pipeline, which deterministically generates QA pairs from pre-existing human-annotated semantic change masks, demonstrates that large-scale temporal QA datasets can be created with minimal additional manual effort once reliable change annotations are available. The accompanying ChronoInstruct~(104,949 samples spanning multiple answer formats) further illustrates how this pipeline can be extended to produce instruction-tuning data at scale. We will release the construction code alongside the datasets, and we hope this transparent and reproducible workflow can lower the barrier for future dataset development in the temporal remote sensing domain.

\paragraph{Insights into temporal modeling for remote sensing.}
Our extensive benchmarking of both commercial and open-source MLLMs reveals that Long-Term Memory is the most critical bottleneck in current models, which may help prioritize future research directions. The GeoChrono model, while still far from human-level performance on the most demanding tasks~(e.g., OHM), provides preliminary validation that explicitly modeling per-location temporal trajectories, leveraging the geostationary prior unique to remote sensing, can yield meaningful improvements. The Coarse-to-Fine Token Compressor further suggests that the inherent spatial redundancy in wide-area remote sensing scenes can be exploited to substantially reduce computational cost without proportional performance degradation. We view these as initial steps rather than definitive solutions, and hope they can serve as useful baselines for the community to build upon.

\paragraph{Open resources.}
To facilitate reproducibility and encourage follow-up research, we will publicly release ChronoBench, ChronoInstruct, the rule-based construction code, the evaluation toolkit, and the trained GeoChrono model weights. We hope these resources collectively support the community in advancing long-term temporal understanding in remote sensing and related Earth observation applications.











\end{document}